\documentclass[pdflatex,sn-mathphys-num]{sn-jnl}

\usepackage{graphicx}%
\usepackage{multirow}%
\usepackage{amsmath,amssymb,amsfonts}%
\usepackage{amsthm}%
\usepackage{mathrsfs}%
\usepackage[title]{appendix}%
\usepackage{xcolor}%
\usepackage{textcomp}%
\usepackage{manyfoot}%
\usepackage{booktabs}%
\usepackage{algorithm}%
\usepackage{algorithmicx}%
\usepackage{algpseudocode}%
\usepackage{listings}%

\usepackage{adjustbox}

\theoremstyle{thmstyleone}%
\theoremstyle{thmstyletwo}%

\theoremstyle{thmstylethree}%
\usepackage{array}
\usepackage{gensymb}
\usepackage{xr} 

\usepackage{xcolor,colortbl}
\definecolor{d}{RGB}{245,220,215}
\definecolor{l}{RGB}{252,240,225}

\begin{document}
\title[Article Title]{
From Promise to Practical Reality: Transforming Diffusion
MRI Analysis with Fast Deep Learning Enhancement
}


\author*[1,2]{\fnm{Xinyi} \sur{Wang}}\email{xwan2191@uni.sydney.edu.au}

\author[2,3,4]{\fnm{Michael} \sur{Barnett}}\email{michael@sydneyneurology.com.au}

\author[5]{\fnm{Frederique} \sur{Boonstra}}\email{frederique.boonstra@monash.edu}

\author[3,6]{\fnm{Yael} \sur{Barnett}}\email{Yael.Barnett@svha.org.au}

\author[2,7]{\fnm{Mariano} \sur{Cabezas}}\email{mariano.cabezas@sydney.edu.au}

\author[2,8]{\fnm{Arkiev} \sur{D'Souza}}\email{arkiev.dsouza@sydney.edu.au}

\author[9,10]{\fnm{Matthew} \sur{C. Kiernan}}\email{matthew.kiernan@neura.edu.au}

\author[2,3]{\fnm{Kain} \sur{Kyle}}\email{Kain@snac.com.au}

\author[11,12]{\fnm{Meng} \sur{Law}}\email{meng.law@monash.edu}

\author[13]{\fnm{Lynette} \sur{Masters}}\email{Lynette.Masters@i-med.com.au}

\author[1,2,3,14]{\fnm{Zihao} \sur{Tang}}\email{zihao.tang@sydney.edu.au}

\author[6]{\fnm{Stephen} \sur{Tisch}}\email{stephen.tisch@svha.org.au}

\author[2,14]{\fnm{Sicong} \sur{Tu}}\email{sicong.tu@sydney.edu.au}

\author[5,15]{\fnm{Anneke} \sur{Van Der Walt}}\email{anneke.vanderwalt@monash.edu}

\author[2,3]{\fnm{Dongang} \sur{Wang}}\email{dongang.wang@sydney.edu.au}

\author[2,8,16]{\fnm{Fernando} \sur{Calamante}}\email{fernando.calamante@sydney.edu.au}
\equalcont{These authors contributed equally to this work.}

\author[1,2]{\fnm{Weidong} \sur{Cai}}\email{tom.cai@sydney.edu.au}
\equalcont{These authors contributed equally to this work.}

\author*[2,3,14]{\fnm{Chenyu} \sur{Wang}}\email{chenyu.wang@sydney.edu.au}
\equalcont{These authors contributed equally to this work.}

\affil[1]{\orgdiv{School of Computer Science}, \orgname{University of Sydney}, \orgaddress{\city{Sydney}, \state{NSW}, \country{Australia}}}

\affil[2]{\orgdiv{Brain and Mind Center}, \orgname{University of Sydney}, \orgaddress{\city{Sydney}, \state{NSW}, \country{Australia}}}

\affil[3]{\orgname{Sydney Neuroimaging Analysis Center}, \orgaddress{\city{Sydney}, \state{NSW}, \country{Australia}}}

\affil[4]{\orgdiv{Department 
of Neurology}, \orgname{Royal Prince Alfred Hospital}, \orgaddress{\city{Sydney}, \state{NSW}, \country{Australia}}}

\affil[5]{\orgdiv{Department of Neuroscience, School of Translational Medicine}, \orgname{Monash University}, \orgaddress{\city{Melbourne}, \state{VIC}, \country{Australia}}}

\affil[6]{\orgdiv{Department
of Radiology}, \orgname{St Vincent's Hospital}, \orgaddress{\city{Sydney}, \state{NSW}, \country{Australia}}}

\affil[7]{\orgdiv{Macquarie University Hearing}, \orgname{Macquarie University}, \orgaddress{\city{Sydney}, \state{NSW}, \country{Australia}}}

\affil[8]{\orgdiv{Sydney Imaging}, \orgname{University of Sydney}, \orgaddress{\city{Sydney}, \state{NSW}, \country{Australia}}}

\affil[9]{\orgdiv{Neuroscience Research Australia}, \orgname{University of New South Wales}, \orgaddress{\city{Sydney}, \state{NSW}, \country{Australia}}}

\affil[10]{\orgdiv{School of Clinical Medicine}, \orgname{University of New South Wales}, \orgaddress{\city{Sydney}, \state{NSW}, \country{Australia}}}

\affil[11]{\orgdiv{Department of Neuroscience, Central Clinical School}, \orgname{Monash University}, \orgaddress{\city{Melbourne}, \state{VIC}, \country{Australia}}}

\affil[12]{\orgdiv{Department of Radiology}, \orgname{Alfred Health}, \orgaddress{\city{Melbourne}, \state{VIC}, \country{Australia}}}

\affil[13]{\orgname{I-MED Radiology}, \orgaddress{\city{Sydney}, \state{NSW}, \country{Australia}}}

\affil[14]{\orgdiv{Sydney Medical School}, \orgname{University of Sydney}, \orgaddress{\city{Sydney}, \state{NSW}, \country{Australia}}}

\affil[15]{\orgdiv{Department of Neurology}, \orgname{The Alfred Hospital}, \orgaddress{\city{Melbourne}, \state{VIC}, \country{Australia}}}


\affil[16]{\orgdiv{School of Biomedical Engineering}, \orgname{University of Sydney}, \orgaddress{\city{Sydney}, \state{NSW}, \country{Australia}}}


\abstract{
Fiber orientation distribution (FOD) is an advanced diffusion MRI modeling technique that represents complex white matter fiber configurations, and a key step for subsequent brain tractography and connectome analysis. Its reliability and accuracy, however, heavily rely on the quality of the MRI acquisition and the subsequent estimation of the FODs at each voxel. Generating reliable FODs from widely available clinical protocols with single-shell and low-angular-resolution acquisitions remains challenging but could potentially be addressed with recent advances in deep learning-based enhancement techniques. Despite advancements, existing methods have predominantly been assessed on healthy subjects, which have proved to be a major hurdle for their clinical adoption. In this work, we validate a newly optimized enhancement framework, FastFOD-Net, across healthy controls and six neurological disorders. 
This accelerated end-to-end deep learning framework enhancing FODs with superior performance and delivering training/inference efficiency for clinical use ($60\times$ faster comparing to its predecessor).
With the most comprehensive clinical evaluation to date, our work demonstrates the potential of FastFOD-Net in accelerating clinical neuroscience research, empowering diffusion MRI analysis for disease differentiation, improving interpretability in connectome applications, and reducing measurement errors to lower sample size requirements. Critically, this work will facilitate the more widespread adoption of, and build clinical trust in, deep learning based methods for diffusion MRI enhancement. Specifically, FastFOD-Net enables robust analysis of real-world, clinical diffusion MRI data, comparable to that achievable with high-quality research acquisitions.
}



\maketitle

\section{Introduction}\label{sec1}
Analysis of diffusion-weighted imaging (DWI) can map structural brain connectivity non-invasively. Through constrained spherical deconvolution  (CSD)~\citep{jeurissen2014multi,tournier2007robust,tournier2004direct}, high-quality fiber orientation distribution (FOD) estimates for complex white matter (WM) fiber bundle configurations can be reliably computed from multi-shell (i.e., DWI data acquired with multiple b-values) acquisitions of high-angular-resolution diffusion imaging (HARDI), yielding accurate WM tractography~\citep{calamante2019seven,dell2019modelling}. In clinical research settings, however, single-shell low-angular-resolution diffusion imaging (LARDI) is more commonly acquired due to the logistical constraints imposed on non-research-focused clinical sites, particularly, limited acquisition times for clinical patients~\cite{zeng2022fod}. 
This results in a substantial discrepancy in quality between high-quality research-grade protocols and those used in clinical practice. 
Additionally, the bulk of historical research datasets often lack the multi-shell diffusion acquisitions necessary for advanced analysis~\citep{mueller2005ways,marek2011parkinson}. Consequently, clinical and historical research diffusion MRI data often lacks sufficient information to reliably reconstruct complex fiber structures, which may lead to misleading analysis outputs~\citep{farquharson2013white}. These challenges emphasize the need for enhancement methods that can produce reliable results from data acquired with non-state-of-the-art research diffusion protocols.

Deep learning has been introduced recently to bridge the gap between the need for advanced research protocols and the constrained settings of clinical diffusion MRI protocols. One of these recent methods directly enhanced single FOD estimates obtained from patches of minimally pre-processed single-shell, clinical-quality data, achieving enhancements comparable to those from multi-shell HARDI acquisitions~\citep{zeng2022fod}. 
The few other related attempts~\citep{lucena2021enhancing,rauland2023using,bartlett2024recovering,da2024fod,wang2025ufree} have only been validated on small-scale datasets of healthy subjects and FOD-level metrics (i.e., an intermediate step in the analysis pipeline), limiting practical adoption in disease-focused clinical research studies.

\begin{figure*}
        \centering
        \includegraphics[width=\linewidth]{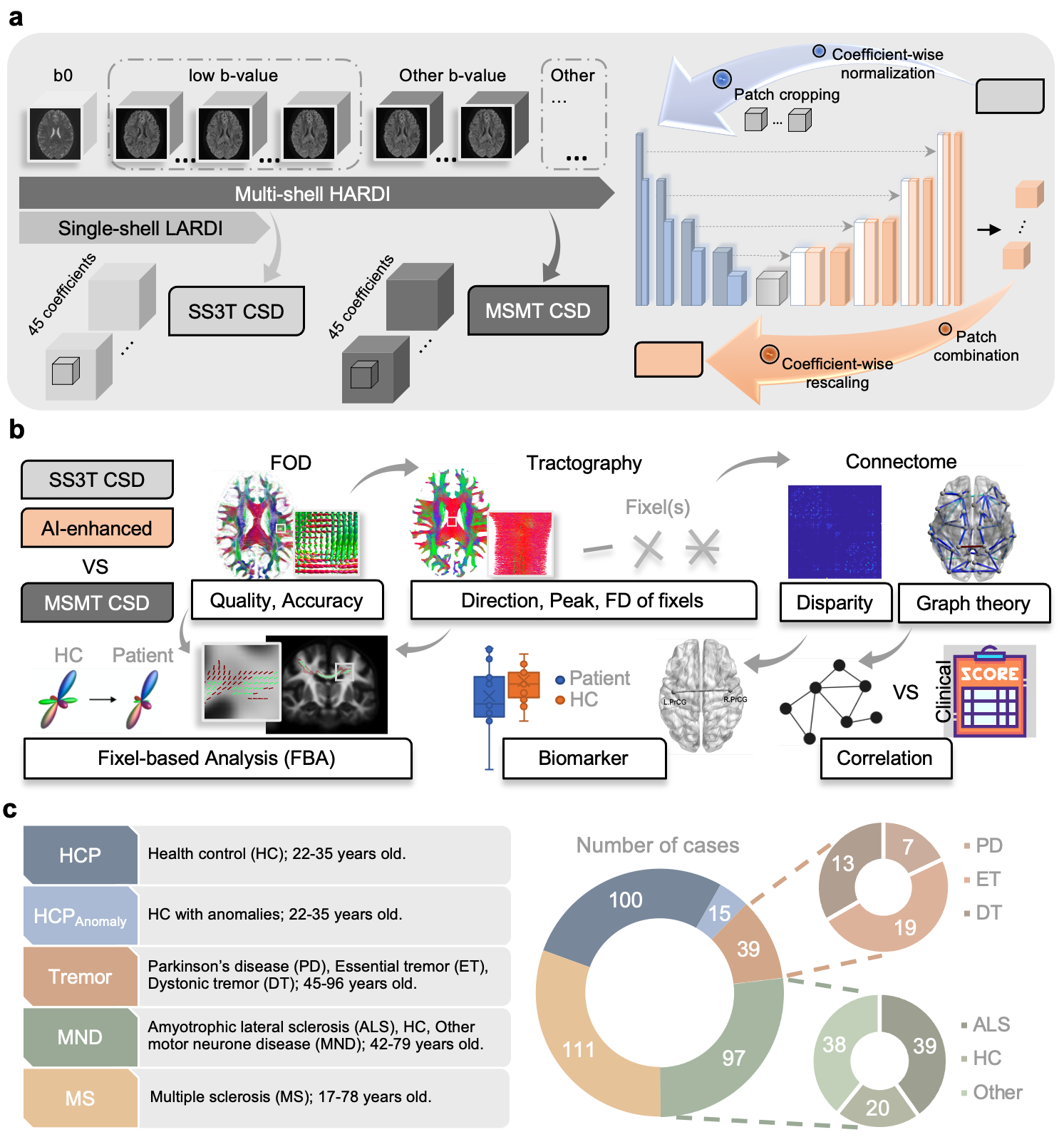}
        \caption{\textbf{Overview of analysis workflow and data composition.} \textbf{a,} Left: Illustration of data preparation. Multi-shell high-angular-resolution diffusion imaging (HARDI) data were acquired at multiple b-values with a large number of diffusion gradient directions (typically over 60 directions)~\citep{zeng2022fod}, while low-angular-resolution diffusion imaging (LARDI) data typically combine a single b0 image with a smaller set of directions, often from a single low b-value acquisition. FOD coefficients were generated using single-shell 3-tissue (SS3T) constrained spherical deconvolution (CSD)~\citep{dhollander2016unsupervised} for LARDI (input, light gray) and multi-shell multi-tissue (MSMT) CSD~\citep{jeurissen2014multi} for HARDI, serving as the ground truth (GT, dark gray). Right: We present FastFOD-Net, a fast deep learning method with an encoder (blue)–decoder (orange) architecture that efficiently estimates high-quality FOD coefficients (orange rectangle) from SS3T CSD data (gray rectangle). 
        \textbf{b,} The evaluation guideline contains diverse end-point diffusion analysis metrics to fully investigate the potential of deep learning solutions in clinical neuroscience research. 
        \textbf{c,} Composition of data used in this study: the Human Connectome Project (HCP), HCP subjects with MRI anomalies (HCP$_{Anomaly}$), and datasets (denoted as Tremor, MND, and MS) from patients with tremors, motor neuron diseases, and multiple sclerosis.}
\label{fig:overview}
\end{figure*}

In this article, we establish an evaluation framework for learning-based FOD enhancement (Fig.~\ref{fig:overview}) to compare and clinically validate a newly optimized variant, FastFOD-Net. By introducing patch-wise training and inference, instead of time-consuming voxel-wise predictions, the new FastFOD-Net algorithm obtains superior performance while greatly improving the computational efficiency, by at least 20 times for training and 60 times for inference, compared to its predecessor~\cite{zeng2022fod}. We conducted a comprehensive clinical validation through comparison with state-of-the-art diffusion analysis methods, including fixel-based analysis (FBA)~\cite{dhollander2021fixel} and connectome analysis~\cite{civier2019removal}, to fully investigate the potential of the proposed enhancement framework. The clinical validation presented in this work encompasses applications not only in healthy controls but also in datasets from diverse neurological conditions, including Parkinson’s disease (PD), essential tremor (ET), dystonic tremor (DT), amyotrophic lateral sclerosis (ALS), other motor neuron diseases, and multiple sclerosis (MS). 
Our results highlight the reliability of using FastFOD-Net to bridge the gap between advanced research protocols and constrained clinical settings; and will facilitate the wider adoption of deep learning enhancement solutions to advance clinical neuroscience research using diffusion MRI data. The project page of FastFOD-Net, along with our evaluation framework, is available at: \url{https://github.com/MAGNOLIAw/FastFOD-Net}.

\section{Results}\label{sec:results}
The experiments were conducted on datasets from the Human Connectome Project (HCP and HCP$_{Anomaly}$)~\citep{van2013wu}, Tremor~\citep{kyle2023tremor}, MND~\citep{almgren2025quantifying}, and MS---see demographics and data details in Fig.~\ref{fig:overview}c, Methods~\ref{sec:data_processing}, and Supplementary Table 1. Four different methods to compute the FOD were compared: single-shell 3-tissue (SS3T) CSD~\cite{dhollander2016unsupervised} and multi-shell multi-tissue (MSMT) CSD~\cite{jeurissen2014multi}, which are the state-of-the-art for analyzing single- and multi-shell DWI data, respectively; FOD-Net~\citep{zeng2022fod}, a recent enhancement baseline method; and the new optimized FastFOD-Net method---see Methods~\ref{sec:methods} for details. The enhancement performance was evaluated by comparing pre-enhancement (SS3T CSD), post-enhancement (FOD-Net/FastFOD-Net), with the ground truth (GT, MSMT CSD).

\subsection{FOD assessment}
The quantitative FOD assessments evaluated the signal-to-noise and angular accuracy in the following brain regions as described previously~\citep{zeng2022fod}: (1) deep WM regions only, and (2) WM/gray matter (GM) border area including a mixture of WM and GM at juxtacortical (JCWM) or subcortical regions (WM-SGM). Fig.~\ref{fig:fodfixel_barboxplots}a,b show that the proposed FastFOD-Net is significantly superior across all conditions for both FOD-level metrics (impact of noise and angular accuracy)---see details on metrics and results in Methods~\ref{sec:fod_assessment} and Supplementary Table 3.

\subsection{Fiber bundle element assessment}
\begin{figure*}
        \centering
        \includegraphics[width=\linewidth]{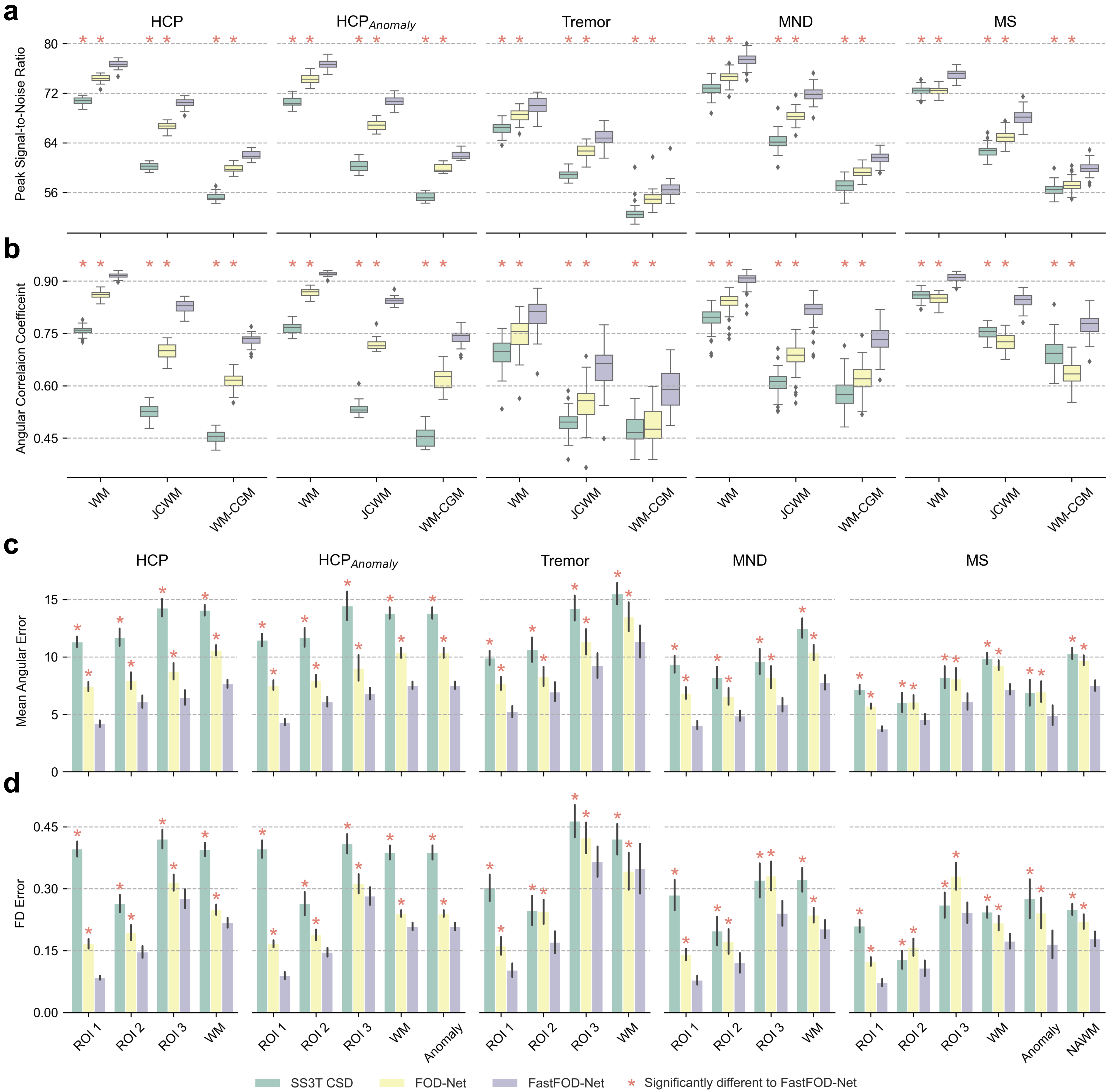}
        \caption{\textbf{Quantitative analysis of FOD and fixel assessments.} Error bars represent standard deviation, and * denotes results are significantly different from FastFOD-Net ($p < 0.05$). 
        \textbf{a,} Box plots of peak signal-to-noise ratio ($\mathbf{PSNR}$) between the ground truth (GT, corresponding to MSMT CSD) and other methods (SS3T CSD, FOD-Net, and FastFOD-Net) across patients and controls. WM denotes white matter; JCWM and WM-SGM represent the boundaries between WM and juxtacortical or subcortical gray matter. 
        \textbf{b,} Box plots of angular correlation coefficient ($\mathbf{r_{Angular}}$) between GT and other methods. \textbf{c,} Bar plots of mean angular error ($\mathbf{\mu E_{Angular}}$) between GT fixels and other estimations across patients and controls. ROI 1, 2, and 3 represent areas with distinct fiber configurations: 1 single fiber population, 2-way fibre crossing, and 3-way fiber crossing, respectively. ‘Anomaly’ represents ROIs surrounding the anomalous regions on HCP$_{Anomaly}$ cases, and lesions on MS cases. NAWM represents normal-appearing WM outside lesions. \textbf{d,} Bar plots of fiber density (FD) error ($\mathbf{E_{FD}}$) between GT fixels and other estimations across patients and controls.}
\label{fig:fodfixel_barboxplots}
\end{figure*}

The FOD data can therefore be discretized in the angular domain based on its fixels, where `fixel' refers to a specific fiber bundle population within a voxel~\cite{dhollander2021fixel}. The impact of the FOD enhancement can then be assessed using fixel-wise metrics, focusing on the angular and amplitude properties of the FOD lobes---see Methods~\ref{sec:methods_fixel} for details. Fig.~\ref{fig:fodfixel_barboxplots}c,d demonstrates that FastFOD-Net significantly ($p < 0.05$) outperformed SS3T CSD and FOD-Net in terms of mean angular error $\mathbf{\mu E_{Angular}}$ and fiber density (FD) error $\mathbf{E_{FD}}$ of fiber bundle elements in all regions of interest (ROIs) considered (see Supplementary Methods 1.5 for ROI definitions) and all datasets. Supplementary Fig. 3 shows FastFOD-Net obtained peak errors $\mathbf{E_{Peak}}$ that were comparable to its predecessor FOD-Net. More detailed results are provided in Supplementary Tables 4-6.

\subsection{Connectome assessment}\label{sec:connectome_accuracy_results}
The structural connectome matrix computed from diffusion MRI fiber-tracking data serves as a summary depiction of the whole-brain connectivity, offering insights into circuit-based alterations associated with neurological and psychiatric disorders~\citep{fornito2012schizophrenia,fornito2015connectomics}. Details of the metrics used to reflect the accuracy of the estimated connectomes, when compared to the GT connectome computed with MSMT CSD of multi-shell HARDI data, are provided in Methods \ref{sec:methods_connectome}. 

\begin{figure*}[t]
\centering
    \includegraphics[width=\linewidth]{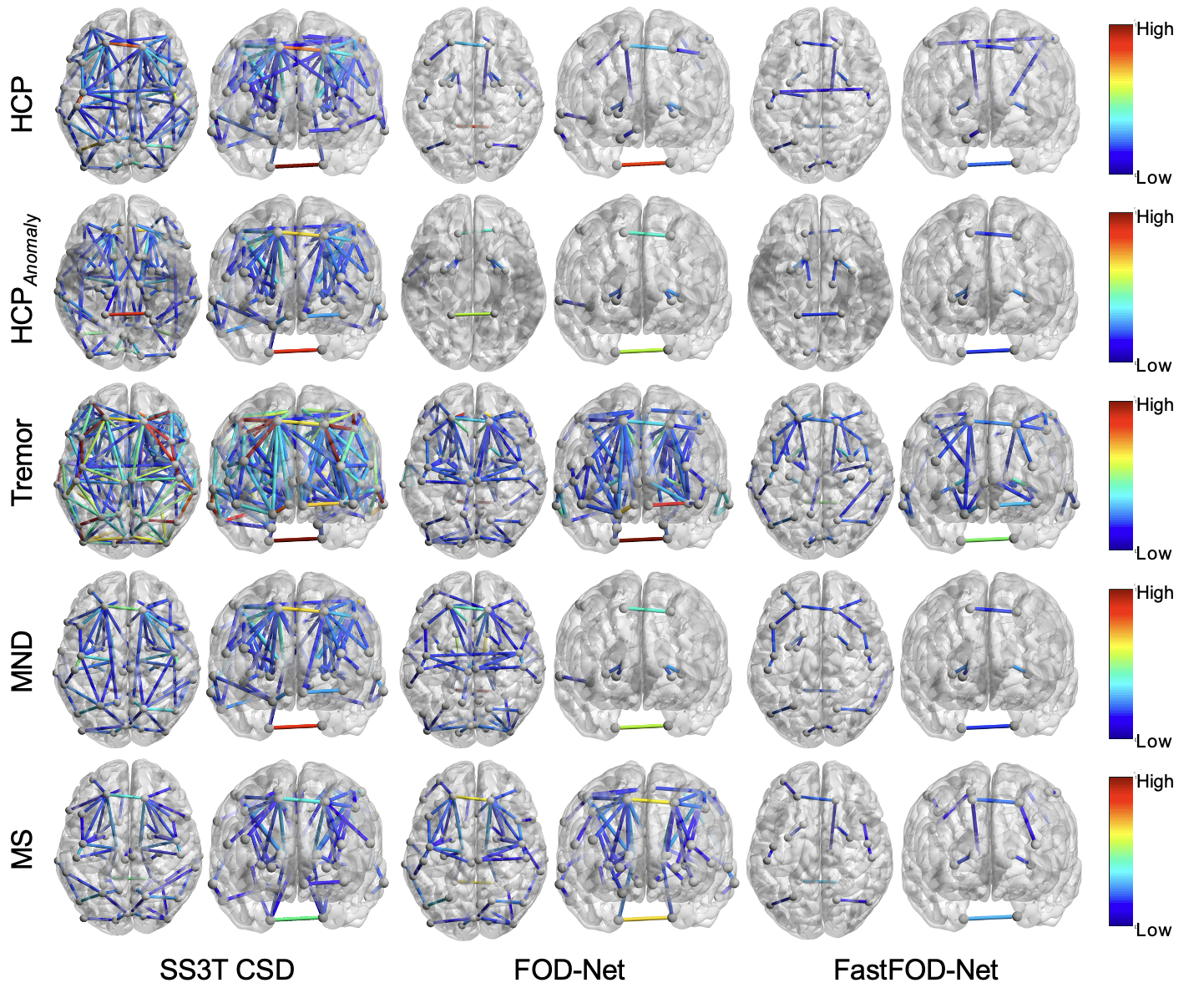}\caption{\textbf{Quantitative analysis of connectome disparity.} Axial and coronal views of mean edge disparity matrices between the ground truth (GT) via MSMT CSD and other LARDI-based estimates via SS3T CSD, FOD-Net, and FastFOD-Net. The color scale ranges from blue to red, representing edges with a gradient from low to high disparity. For sparsity, the visualization threshold is set to 4000.} 
\label{fig:disp_scatterbrainnetplots}
\end{figure*}

FastFOD-Net produced high-quality connectomes, with the lowest disparity (Fig.~\ref{fig:disp_scatterbrainnetplots} and Supplementary Table 7), the fewest significant different edges relative to GT (Supplementary Table 7 and Supplementary Fig. 4), and the highest Kendall correlation (Supplementary Fig. 5 and Supplementary Table 7), surpassing both SS3T CSD and FOD-Net by a substantial margin. To zoom in on the differences between the learning-based methods, FastFOD-Net reduced the disparity of interhemispheric connections (Supplementary Fig. 6 and 7), whereas FOD-Net did not reduce and even exacerbated the disparity in subjects with tremor disease variants and motor neuron disorders. 

\begin{figure*}
        \centering
        \includegraphics[width=\linewidth]{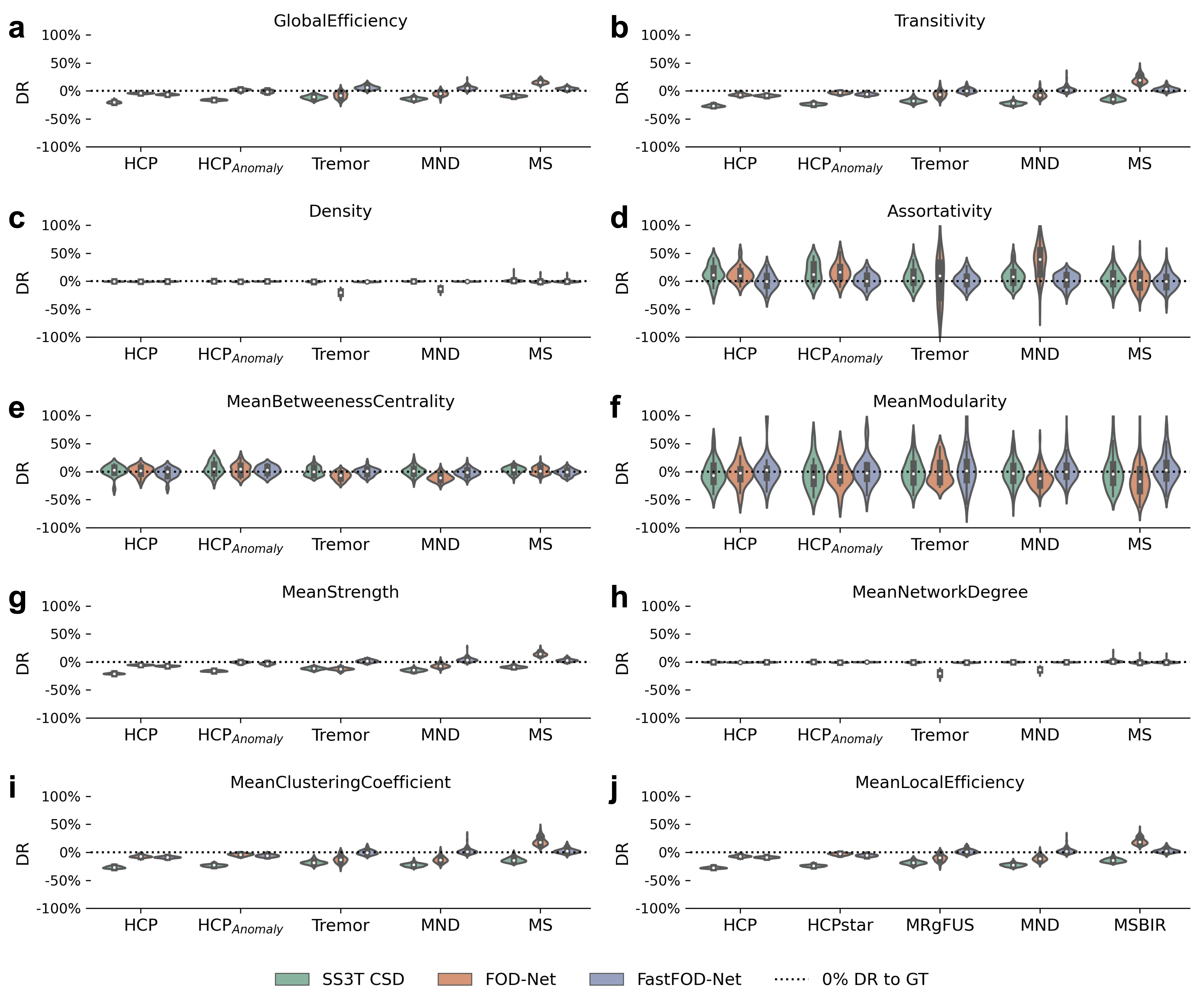}
        \caption{\textbf{Quantitative analysis of graph metrics.} Violin plots of the difference ratio (DR) of graph metrics between MSMT CSD-derived connectomes (ground truth, GT) and other estimations (from SS3T CSD, FOD-Net, and FastFOD-Net) in HCP, HCP$_{Anomaly}$, Tremor, MND, and MS cases.
        The metrics analyzed include: \textbf{a,} global efficiency; \textbf{b,} transitivity; \textbf{c,} density; \textbf{d,} assortativity; \textbf{e,} mean between centrality; \textbf{f,} mean modularity; \textbf{g,} mean strength; \textbf{h,} mean network degree; \textbf{i,} mean clustering coefficient; \textbf{j,} mean local efficiency.
        } 
\label{fig:graphmetrics_scatterviolinplots}
\end{figure*}

Finally, the network topology of the connectome is often summarized with graph-theoretical metrics~\cite{yeh2021mapping}. 
As illustrated in Fig.~\ref{fig:graphmetrics_scatterviolinplots}, the difference ratio (DR) to the GT for FastFOD-Net is the closest to 0\% for most graph measures, confirming its superiority compared to the other LARDI-based methods. Numerical results on the DR of graph metrics are provided in Supplementary Tables 8 and 9.

\begin{figure*}[!h]
    \centering
    \includegraphics[width=\linewidth]{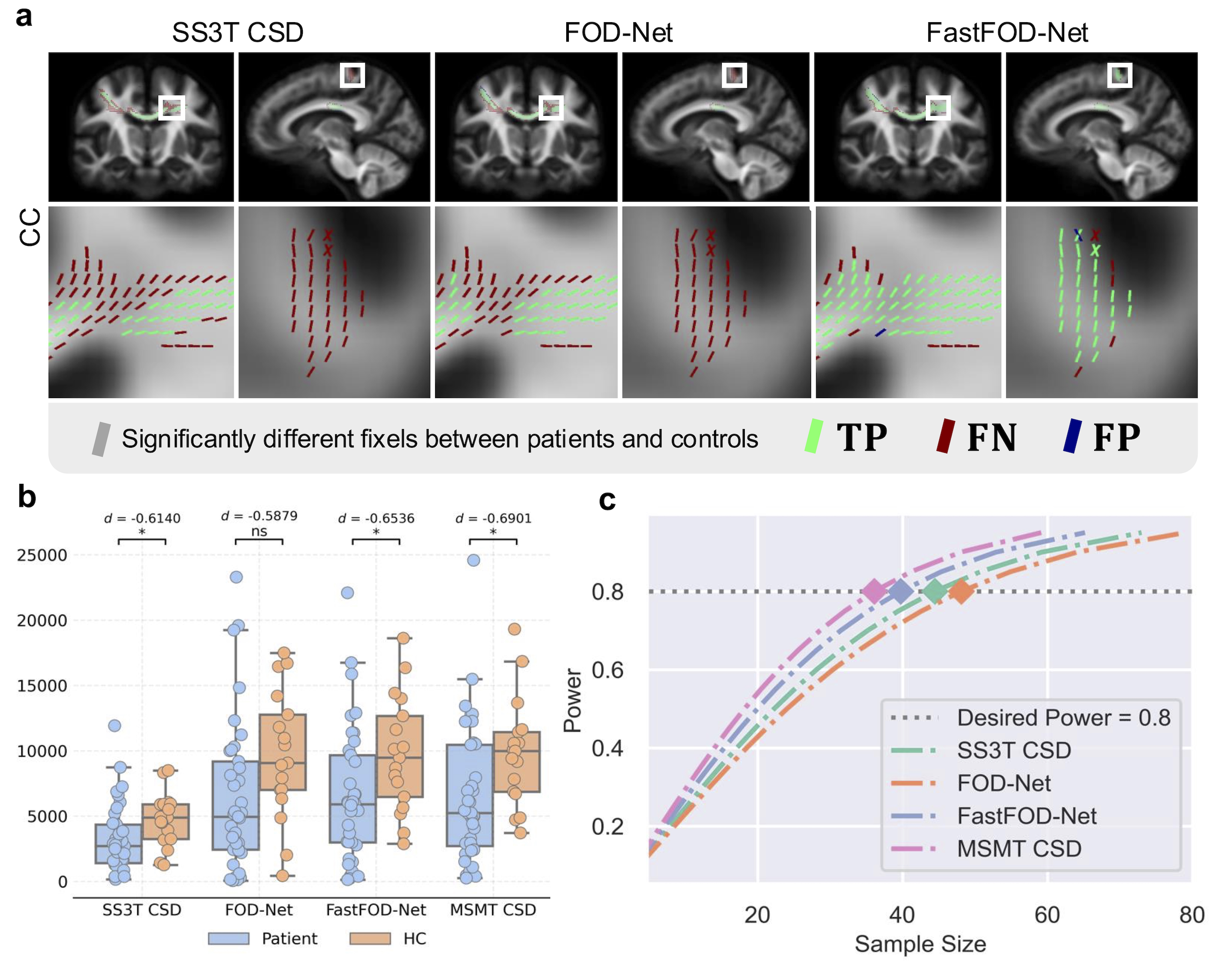}
    \caption{\textbf{Analysis of clinical neuroscience applications in patients with amyotrophic lateral sclerosis (ALS).} \textbf{a,} Comparison of significantly different fixels between the fiber density (FD) of amyotrophic lateral sclerosis (ALS) patients and healthy controls (HCs) using MSMT CSD as the reference ground truth (GT) method---see more details in Supplementary Fig. 8 and Supplementary Table 10. 
    In this comparison, significant fixels derived from the GT method were juxtaposed with those obtained from alternative methodologies (specifically, SS3T CSD, FOD-Net, and FastFOD-Net). Notably, fixels correctly identified ($\mathbf{TP}$) by the comparison methods along the corpus callosum (CC) are denoted in green. Conversely, fixels that were not identified despite being significantly different with MSMT CSD ($\mathbf{FN}$) are depicted in red, while those erroneously identified as significant ($\mathbf{FP}$) are marked in blue. True negative fixels ($\mathbf{TN}$) are omitted for clarity, focusing on detected and real differences. 
    \textbf{b,} Statistical significance analysis of a pathology-relevant connection between ALS patients and controls. The connection of interest, from the left precentral gyrus (L.PrCG) to the right precentral gyrus (R.PrCG), is assessed using independent t-tests. Above each pair of box plots, p-values indicate the statistical significance of group differences ($p < 0.05$ denoted by ‘*’, while `ns' indicates non-significance). Cohen’s d effect sizes further quantify the magnitude of these differences.
    \textbf{c,} Sample size requirements to detect true differences of this pathology-relevant connection between ALS patients and HCs.
    }
    \label{fig:applicationsvis}
\end{figure*}

\subsection{Clinical neuroscience applications}\label{sec:study_results}
\noindent\textbf{Fixel-based analysis} 
FBA~\cite{raffelt2015connectivity,raffelt2017investigating} is a powerful tool for performing group comparison analysis at the fixel level (i.e., testing for fixel-wise rather than voxel-wise differences) and has previously unmasked differences between patients with ALS and healthy controls (HCs)~\cite{almgren2025quantifying}. Here, we investigated differences between ALS patients and HCs for each FOD method using the same MRI protocol and scanner, with MSMT CSD findings serving as the GT. Details on the methodology for FBA are provided in Methods~\ref{sec:methods_applications}. 

FBA results from SS3T CSD presented a high number of false negatives $\mathbf{FN}$, i.e., missed fixel differences (Fig.~\ref{fig:applicationsvis}a and Supplementary Fig. 8). As shown in the figures, FastFOD-Net addressed this gap by detecting a larger number of true significant differences, true positives ($\mathbf{TP}$), than FOD-Net and SS3T CSD. This improvement was particularly notable in the corpus callosum (CC). Furthermore, FOD-Net produced a higher number of implausibly significant differences, i.e., false positives ($\mathbf{FP}$), in the corticospinal tract (CST), as observed in the coronal view. In Supplementary Table 10, FastFOD-Net achieved a $\sim20\%$ enhancement in both $\mathbf{Sensitivity}$ and $\mathbf{F1\ score}$ along the CC compared to SS3T CSD, whereas FOD-Net did not achieve a similar level of improvement. When looking at the CST, both learning-based methods increased $\mathbf{F1}$ by $\sim10\%$. This has direct implications for diagnosis and disease monitoring in ALS as the CC and CST are the two most salient discriminant disease features in ALS~\cite{almgren2025quantifying,tu2020regional}.
\\

\noindent \textbf{Pathological connection analysis}
For pathology-related connections, alterations in connectivity values between clinical groups were analyzed to further explore the practical impact of learning-based FOD enhancement in a real-world clinical study (Methods~\ref{sec:methods_applications}). 

Fig~\ref{fig:applicationsvis}b demonstrates that FastFOD-Net reconstructs connectomes that better match the reference MSMT CSD in the ALS patient cohort. On the one hand, similar to MSMT CSD, FastFOD-Net revealed significantly ($p < 0.05$) reduced connectivity in ALS (left precentral gyrus [L.PrCG] to right precentral gyrus [R.PrCG]), consistent with previous reports~\cite{rose2012direct}, whereas its predecessor FOD-Net failed to detect this. We further evaluated the sample size required to separate patients and HC groups based on the connection from the primary motor cortex when a desired power of 0.8 is achieved across all methods (Fig~\ref{fig:applicationsvis}c). FOD-Net required slightly more samples than SS3T CSD, while FastFOD-Net performed better than both, suggesting that true pathological connections were successfully recovered and analyzed. MSMT CSD required the fewest samples, followed by FastFOD-Net, which may indicate an optimization of research effort.
For the connection between the left postcentral gyrus (L.PoCG) to the right postcentral gyrus (R.PoCG), all methods correctly showed non-significant differences, in alignment with previous findings~\cite{rose2012direct}. However, SS3T CSD, yielded the largest absolute effect size of 0.54. In contrast, FastFOD-Net demonstrated superior reliability by reducing the likelihood of false-positives findings (absolute effect size of 0.25) (Supplementary Fig. 9).

The tremor movement disorder cohorts lacked matched healthy control data, no significant differences between the different disease groups were detected by MSMT CSD or SS3T CSD at either the fixel or connectome level. Consequently, further biomarker analysis was not performed even though the enhancement also showed no significant differences as expected (thus avoid false-positive connections)---see details of results in Supplementary Table 11.
\\

\noindent \textbf{Clinical correlation analysis}
The capability for reconstructing white matter tracts from focal lesions has been raised as one of the challenges for tractography in diseases such as multiple sclerosis, where the heterogeneity of lesion load and distribution may lead to false connections and biased outcomes. 

We assessed the correlation between graph metrics derived from various FOD reconstructions and both conventional imaging and clinical outcome, including whole-brain atrophy, disease duration, and Expanded Disability Status Scale (EDSS) within 12 months. The FastFOD-Net approach demonstrated significantly higher correlations with lesion volume and disease duration in both mean betweenness centrality and mean clustering coefficient, aligning with previously reports~\cite{abad2015analysis}. In contrast, SS3T CSD did not show significant correlations with lesion volume and disease duration in mean betweenness centrality and exhibited lower correlation values in mean clustering coefficient. Several graph metrics, including mean strength, mean transitivity, and mean global efficiency, showed statistically significant relationships with EDSS using FastFOD-Net~\cite{pagani2020structural,shu2011diffusion,kocevar2016graph}. Consistent with prior findings, FastFOD-Net captured significant relationships with EDSS in these graph metrics, similar to MSMT CSD. Assortativity, however, did not show a significant correlation with EDSS~\cite{pagani2020structural}, and FastFOD-Net, like MSMT CSD, avoided false-positive correlations in this metric.
We assessed the correlation between graph metrics derived from various FOD reconstructions and both conventional imaging and clinical outcome, including whole-brain atrophy, disease duration, and Expanded Disability Status Scale (EDSS) score, captured within 12 months. The FastFOD-Net approach demonstrated significantly higher correlations with lesion volume and disease duration in both mean betweenness centrality and mean clustering coefficient, aligning with previously reports~\cite{abad2015analysis}. In contrast, SS3T CSD did not show significant correlations with lesion volume and disease duration in mean betweenness centrality and exhibited lower correlation values in mean clustering coefficient. Several graph metrics, including mean strength, mean transitivity, and mean global efficiency, showed statistically significant relationships with EDSS using FastFOD-Net~\cite{pagani2020structural,shu2011diffusion,kocevar2016graph}. Consistent with prior findings, FastFOD-Net captured significant relationships with EDSS in these graph metrics, similar to MSMT CSD. Assortativity, however, did not show a significant correlation with EDSS~\cite{pagani2020structural}, and FastFOD-Net, like MSMT CSD, avoided false-positive correlations in this metric.

\section{Discussion}\label{sec:discussion}
While there have been major research advances in methods to exploit the wealth of information available from diffusion MRI, their reliability based on clinical-quality imaging data remains a practical challenge. Recent studies in DWI and FOD enhancement~\citep{lyon2022angular,lyon2023spatio,lucena2021enhancing,zeng2022fod,rauland2023using,da2024fod,wang2025ufree,ordinola2025super} have introduced novel learning-based approaches to address this issue, yet have largely neglected the translation of enhanced image quality into real-world clinical advantage. Indeed, the translation of deep learning advances into real-world applications, especially in the clinical sphere,  remains stubbornly rare---despite demonstrable improvements in accuracy and model complexity.
To address this gap, we developed a clinical-trust evaluation framework to assess the real-world, practical impact of learning-based diffusion MRI enhancement. This pipeline adopts commonly used FOD-level metrics and refines fixel-level measures by considering missing and extra fiber elements within a voxel, to more accurately measure the differences in complex fiber configurations between different methods. Furthermore, a comprehensive connectome assessment was incorporated to evaluate metrics at the connectome level and topological differences between connectomes reconstructed from different FOD estimations (graph level). This extensive battery of tests, applied to a range of neurological conditions, provides a collection of missing evidence needed for the widespread adoption of deep learning solutions for FOD enhancement in clinical research applications.

Since most prior DWI enhancement methods considered a varying number of gradient directions, their computational cost was relatively high (often more than a week with multiple GPUs) due to the use of recurrent CNNs (RCNNs) and the replacement of convolutions with subsampled multilayer perceptrons (MLPs)~\citep{lyon2022angular,lyon2023spatio}. Using FODs as an intermediate model may be a better solution, as it is more closely related to downstream tasks of interest (e.g., fiber-tracking, FBA, connectome analysis, etc.). Earlier studies on FOD enhancement~\citep{lucena2021enhancing,zeng2022fod,rauland2023using,bartlett2024recovering,da2024fod,wang2025ufree,ordinola2025super} utilized similar deep learning methods, but were mostly evaluated on a limited number of healthy subjects and a restricted set of metrics. From these methods, we chose FOD-Net~\citep{zeng2022fod} as a representative method that has been reported to have good performance. 
During training and inference, enhanced spherical harmonic coefficients were predicted voxel by voxel for each spherical harmonic order. Despite its benefits, FOD-Net is a computationally expensive method that overlooks the contextual properties of voxels in 3D imaging data and inter-order coefficient relationships. In contrast, by avoiding voxel-wise learning and eliminating computationally intensive architectural components for independent order learning, the proposed FastFOD-Net employs a fully convolutional architecture to predict entire input patches at once, achieving improved performance and faster inference (within a few minutes, compared to a few hours seen using FOD-Net). Thus, FastFOD-Net serves as a simplified end-to-end solution for rapid FOD enhancement, further improving feasibility for large-scale clinical evaluation and adoption.

At the FOD, fixel, and connectome levels, FastFOD-Net was found to provide superior FOD enhancement from clinical single-shell acquisitions across a range of different neurological conditions, including motor neuron diseases, tremor disorders, and multiple sclerosis. With a single GPU, the training and testing time was greatly reduced from 4-8 weeks and 23 hours, respectively (for FOD-Net), to within 1 day and 2-3 minutes, respectively (for FastFOD-Net). When compared to the state-of-the-art SS3T CSD, FOD-Net was superior for healthy subjects; however, it led to a higher number of significantly different edges (relative to the GT values) than SS3T CSD in both Tremor and MND cases (Supplementary Table 7). This limitation was less pronounced in FOD and fixel metrics (Fig.~\ref{fig:fodfixel_barboxplots}). In Fig.~\ref{fig:disp_scatterbrainnetplots}, SS3T CSD and FOD-Net exhibited fewer edges with large disparity, suggesting that they may generate implausible connections that, despite low disparity, deviate significantly from the GT. Many of these connections were absent or near zero in the GT but appeared with nontrivial values, indicating a tendency toward false positives.
In contrast, FastFOD-Net nearly halved the disparity analysis metrics compared to SS3T CSD for disease states, indicating superior robustness in downstream tasks. Similar trends were observed in topological analyses of the connectomes. Overall, the combined analysis of these metrics supports the role of FastFOD-Net for clinical applications to recover information missing from LARDI acquisitions.

ALS is a rare neurodegenerative disorder clinically characterized by a progressive breakdown of essential everyday motor function, ultimately leading to a loss of life~\citep{tu2023amyotrophic}. Electrophysiology studies have consistently identified loss of cortical motor inhibition, i.e., cortical hyperexcitability, as a prominent early marker of disease~\cite{van2019pathophysiology}. On neuroimaging, disease pathology is characterized by clear structural, functional, and metabolic changes~\cite{tu2024pathological}. Diffusivity abnormalities in the CC and CST hold the greatest discriminative accuracy and significantly correlates with severity of functional motor impairment in patients~\cite{almgren2025quantifying,caga2021apathy}. Notably, fine-grained changes within the CST and CC have been shown to reflect clinically meaningful phenotypic variation in asymmetry of disease presentation~\cite{tu2020regional,huynh2021mills}. Our results (Fig.~\ref{fig:applicationsvis}a, Supplementary Fig. 8, and Supplementary Table 10) demonstrated that FastFOD-Net effectively reproduces the detection of most fixels (over 80\%) that significantly differ between ALS and HC along the CC tract when compared with GT. Consequently, FastFOD-Net therefore provides an avenue to investigate the pathophysiology of ALS using routinely collected LARDI diffusion data.

Analysis of the ALS pathology-relevant connections (from L.PrCG to R.PrCG) found that connectomes derived from MSMT CSD (GT) showed higher variability (larger standard deviation) v.s. SS3T CSD ones ( Fig.~\ref{fig:applicationsvis}b), potentially reflecting more faithful recovery of complex white matter fibre configurations along L.PrCG-R.PrCG connectome with the GT acquisition. Since FODs generated with FOD-Net tend to have noisy coefficients with a high number of lobes, FOD-Net demonstrated the largest variability with the widest and most overlapping distributions. Conversely, the GT distribution of this pathological connection better approximated by FastFOD-Net, reducing the likelihood of false positive disease-related differences in clinical-quality diffusion MRI data identified in previous clinical studies~\cite{rose2012direct,baumer2014advances}. The ability of FastFOD-Net to achieve sufficient statistical power with a smaller sample size (Fig.~\ref{fig:applicationsvis}c) reduces the recruitment burden investigate neurological disorders based on clinical-quality imaging data.  

Although FastFOD-Net consistently improved FOD quality across the tremor movement disorder cohorts, small sample sizes (ET: $n=19$, PD: $n=7$, DT: $n=13$) and the absence of healthy controls with a consistent MRI protocol limits the statistical power of group comparisons and the interpretation of disease-specific features. Previou studies have established that each of these disorders exhibit structural connectome changes relative to healthy individuals~\cite{sharifi2014neuroimaging,kamagata2018connectome,bedard2022multimodal}. Bergamino et al.~\cite{bergamino2023white} used ET ($n=17$) and PD ($n=68$); however, direct comparisons among ET, PD, and DT are scarce in the literature, and unique diffusion signatures that distinguish these movement disorders are yet to be defined.
Within this context, our analysis detected no significant differences in connectivity graph metrics between the ET, PD, and DT groups using FastFOD-Net, consistent with the GT MSMT CSD results. FastFOD-Net may potentially preserve true biological networks without introducing spurious group differences, though the limited sample size warrants cautious interpretation. Consistency between FastFOD-Net and MSMT CSD results increases confidence that FastFOD-Net’s enhancements are faithful and do not distort group comparisons. Future studies with larger cohorts and matched healthy controls will be valuable in confirming subtle between-group differences (or the lack thereof) in fixel-based and graph-based metrics.

In MS, FastFOD-Net’s utility for clinical applications was confirmed in a large patient cohort ($n = 111$). MS is an inflammatory and neurodegenerative disease characterized by multifocal white matter lesions and diffuse microstructural damage throughout the brain. Conventional MRI (T2-FLAIR, T1) is incorporated into the diagnostic criteria for MS, facilitating the demonstration of lesions in the CNS that are disseminated in space and time~\cite{thompson2018diagnosis}. However, these sequences are  less sensitive to subtler damage in the brain; and do not reveal changes associated with disease progression independent of relapse biology (i.e., new or enlarging lesions)~\cite{caranova2023systematic}. Additionally, a particular challenge for diffusion-based tractography in MS is the presence of focal lesions that disrupt fiber organization: demyelination and axonal loss lead to low anisotropy, which may impact fiber orientation reconstructions~\cite{lipp2020tractography}. The heterogeneous distribution and variable size of MS lesions make it difficult to obtain consistent fiber orientation information across the brain, posing a stringent test for FOD enhancement methods.
FastFOD-Net performed robustly in our MS cohort, producing high-quality FODs both inside and outside lesions (Fig.~\ref{fig:fodfixel_barboxplots} and Supplementary Fig. 3). This suggests that the deep learning model generalized well to focal tissue damage. Additionally, FastFOD-Net’s FOD enhancement led to subsequent connectome metrics that were biologically meaningful and consistent with those obtained using the gold-standard MSMT CSD approach (Supplementary Fig. 10). For instance, many global network measures (e.g., efficiency, strength, network degree, etc.) computed from FastFOD-Net data showed significant correlations with conventional markers of disease burden, such as total lesion volume, as well as with EDSS (but not whole-brain atrophy). Our findings recapitulate the results of prior studies linking diffusion MRI metrics to MS disease severity. For example, global structural network disruption (e.g., lower network efficiency or density) has been associated with increased disability in people with MS~\cite{charalambous2019structural}. These correlations were highly consistent with those derived from MSMT CSD-based connectomes, suggesting that little information was lost when using FastFOD-Net with SS3T; and that FastFOD-Net retains sensitivity to clinically relevant microstructural changes. This robustness across a large MS sample highlights the potential of FastFOD-Net longitudinal studies and therapeutic trials in MS, where single-shell dMRI could be efficiently acquired while still enabling sophisticated fixel-based analyses or connectomics to monitor disease progression.

Our study has several limitations. First, analysis of the movement disorder cohorts was limited by small sample sizes and the lack of a relevant control group, reducing our ability to detect subtle group differences. Future studies should include larger cohorts of ET, PD, and DT patients, as well as healthy controls, to validate whether FastFOD-Net can reveal network alterations specific to each condition. Additionally, extending FastFOD-Net to other clinical populations and scanner settings would be valuable. MS lesions present a challenging task in white matter for FOD enhancement, but other neurological diseases (such as stroke or traumatic brain injury) exhibit different patterns of white matter damage that warrant evaluation with FOD enhancement. Finally, ongoing improvements in network architecture (e.g., incorporating transformer-based models~\cite{da2024fod}, Mamba-based models~\cite{gu2023mamba}, vector-quantized variational autoencoders [VQ-VAE]~\cite{wang2025ufree}, and diffusion models~\cite{rombach2022high} for FOD super-resolution) may yield even better performance and introduce variability in generations. However, recent benchmarks~\cite{isensee2024nnu,isensee2021nnu,barnett2023real,goebl2025enabling} caution that architectural novelty, on its own, does not guarantee meaningful progress; equally crucial is that large-scale, rigorously validated clinical benefit can also be demonstrated. We anticipate that continued refinement of FOD enhancement methods, combined with thorough clinical validation, will solidify their role in precision neurology—enabling robust connectome and fixel analyses from standard diffusion acquisitions to inform diagnosis, prognosis, and treatment planning.

In conclusion, FastFOD-Net is a fast deep learning method to enhance single-shell LARDI acquisitions within a few minutes, and evaluation across a range of neurological conditions illustrates the potential impact promised of our tool in clinical applications. The results demonstrated FastFOD-Net’s ability to recover fiber information missing from single-shell clinical-quality data, and its superiority for robust downstream brain connectivity analysis. FastFOD-Net's use of single-shell LARDI data (lower acquisition time) and reduced sample size requirements optimize clinical research efforts, while still recovering clinically significant group differences that are the most closely aligned to multi-shell HARDI-quality data without introducing false positives. 

\section{Methods}\label{sec:methods}
\subsection{Datasets and data processing}\label{sec:datasets}
As detailed in Fig.~\ref{fig:overview} and Supplementary Methods 1.1, a total of 362 subjects were included in this work, comprising healthy control subjects and patients with various neurological diseases. We utilized 80 training and 20 validation samples from the Human Connectome Project (HCP)~\citep{van2013wu}, along with an additional 15 validation samples drawn from HCP anomaly cases (HCP$_{Anomaly}$), which comprise datasets with incidental brain abnormalities. Furthermore, we conducted five-fold cross-validation using in-house datasets (denoted as Tremor, MND, MS) from patients with tremor~\citep{kyle2023tremor}, motor neuron diseases, and multiple sclerosis. 
All these datasets were acquired using different high-angular-resolution diffusion imaging (HARDI) protocols, which we selected as the ground truth (GT) data. Details of acquisition parameters are provided in Supplementary Table 1 and Supplementary Methods 1.1, with data processing steps outlined in Supplementary Methods 1.2. Following acquisition and preprocessing of multi-shell HARDI diffusion-weighted imaging (DWI) data and its single-shell clinical-quality counterparts, fiber orientation distribution (FOD) maps were generated with MRtrix~\citep{tournier2012mrtrix,tournier2019mrtrix3} by adopting single-shell three-tissue (SS3T) constrained spherical deconvolution (CSD)~\citep{dhollander2016unsupervised} on single-shell low-angular-resolution diffusion imaging (LADRI) data and multi-shell multi-tissue (MSMT)~\citep{jeurissen2014multi} CSD on the multi-shell HARDI data, respectively. 

\subsection{Architecture}\label{sec:data_processing}
We extended the FOD-Net design~\citep{zeng2022fod} to directly predict entire patches by adopting a fully convolutional backbone~\citep{cciccek20163d}. This conserves computational resources, allowing for greater practical applicability. More details of the architecture and implementation are provided in Supplementary Methods 1.3.

\subsection{Quantitative diffusion MRI assessment}
We presented a clinical-trust evaluation guideline following the analysis workflows of current state-of-the-art diffusion analysis techniques.

\subsubsection{Overall FOD assessment}\label{sec:fod_assessment}
The quantitative FOD assessment was designed to determine whether the gray matter (GM) partial volume influenced the performance of FOD enhancement methods from the FOD perspective, rather than through subsequent derivative measures. Peak signal-to-noise ratio ($\mathbf{PSNR}$) quantified the impact of enhancement on image quality by evaluating the similarity of FODs between ground truth (GT, MSMT CSD) and single-shell based methods (SS3T CSD, FOD-Net, and FastFOD-Net). Additionally, the angular correlation coefficient ($\mathbf{r_{Angular}}$)~\citep{anderson2005measurement} was used to provide a measurement of how correlated the spherical harmonic coefficients are between GT data and other estimates, as formulated in Supplementary Method 1.4. 

\subsubsection{Fiber bundle element assessment}\label{sec:methods_fixel}
Continuous FOD data were segmented into discrete fiber elements using MRtrix~\cite{tournier2012mrtrix,tournier2019mrtrix3}, referred to as `fixels', which represent distinct fiber populations within each voxel. LARDI methods were compared to GT with three key metrics that assess specific fixels within a voxel: mean angular error ($\mathbf{\mu E_{Angular}}$), peak error ($\mathbf{E_{Peak}}$), and fiber density error ($\mathbf{E_{FD}}$)~\cite{zeng2022fod}.
Although the original FOD-Net paper quantified the accuracy of fiber bundle element reconstruction, the effect of unmatched fiber elements between GT fixels and estimated fixels was not considered when computing these measures. Specifically, each fiber bundle element of LARDI estimations was always matched to a GT fixel regardless of their angular error. This implies that matched fiber bundle elements could have perpendicular orientations in a worst-case scenario, leading to misleading results on misaligned fixel pairs. False positive fixels would be ignored even if they were erroneous.
In that sense, $\mathbf{\mu E_{Angular}}$, stands as one of the most valuable metrics because the accurate reconstruction of fiber orientations is key for understanding brain connectivity. 
To fairly compare fiber bundle elements between GT and other methods, estimated fixels with angular errors higher than a selected threshold ($45\degree$) are considered misaligned for the remainder of the paper. Following our definitions of matched and misaligned fixels, we optimized the $\mathbf{E_{Peak}}$ and $\mathbf{E_{FD}}$ measures by also considering misaligned fixels as erroneous predictions. More details of refined fixel metrics are provided in Supplementary Methods 1.5.

Then, we computed these improved metrics in three defined ROIs, containing either a predominantly single-fiber population (corpus callosum [CC]), 2-way crossing, or 3-way crossing, to analyze performance under distinct fiber configurations. Anomaly regions denoting ROIs surrounding the anomalous regions on HCP$_{Anomaly}$ cases, and lesions in MS cases, were also analyzed. Moreover, normal-appearing WM outside lesions was evaluated in MS cases to determine whether deep learning methods performed consistently, regardless of the presence of focal pathology.

\subsubsection{Connectome assessment}~\label{sec:methods_connectome}

The structural connectome~\citep{yeh2019connectomes}, a comprehensive representation of brain connections, was generated using anatomically-constrained tractography (ACT)~\citep{smith2012anatomically} with the input FOD image, as detailed in Supplementary Methods 1.2. It serves as a graph matrix with nodes denoting gray matter regions and edges denoting brain connections between them.
In this work, we first defined the disparity matrix as an edge-wise disparity measurement and employed Kendall's ranking coefficient ($\mathbf{\tau}$)~\citep{kendall1938new} to assess the overall difference between a LARDI-based connectome and its corresponding GT matrix. We also counted the percentage of significantly different edges by extending the disparity analysis with edge-wise paired t-tests ($p < 0.05$) and false discovery rate (FDR) correction. More details of these metrics are provided in Supplementary Methods 1.6. 

In addition, to analyze the connectome enhancement quality from a network perspective, several graph theory metrics~\citep{fornito2016fundamentals} were computed via the Brain Connectivity Toolbox (BCT)~\citep{rubinov2010complex} and compared between methods. Finally, we computed the difference ratio (DR) of these graph metrics as a measure of topological similarity between GT and other methods---see details and formulation in Supplementary Methods 1.6.

\subsubsection{Clinical neuroscience applications}\label{sec:methods_applications}

\textbf{Fixel-based analysis} (FBA)~\cite{raffelt2015connectivity,raffelt2017investigating} was a technique to examine alterations in the microstructural and macrostructural properties of WM tracts at the fixel level, most commonly used when comparing patients to control groups. In this study, our hypothesis is that more significant differences can be observed based on enhanced FODs when compared with no enhancement. 

Several studies have conducted direct investigations into the association between structural brain abnormalities and the cognitive and behavioral profiles of relatively small samples in patients with amyotrophic lateral sclerosis (ALS)~\citep{grossman2008impaired,kasper2014microstructural,libon2012deficits,pettit2013executive}. Based on findings from these studies, we performed FBA analysis on FOD estimates in the CC and corticospinal tract (CST), which are sensitive to impairments in ALS populations when compared to healthy controls (Supplementary Fig. 2). Potential improvements generated by the application of learning-based frameworks cwere assessed by comparison with the analysis of multi-shell HARDI FODs (GT), yielding $\mathbf{TP}$, $\mathbf{FN}$, $\mathbf{FP}$, and $\mathbf{TN}$. To summarize these measures, we computed the ${\mathbf{Sensitivity}}$, ${\mathbf{Specificity}}$ and ${\mathbf{Precision}}$ for each enhancement method. $\mathbf{F1}$ was also included significantimbalance between fixels and non-significantly different fixels---see Supplementary Methods 1.7 for more details.
\\

\noindent\textbf{Pathological connection analysis}
For pathology-related connections, alterations in their connectivity values between clinical groups could be analyzed to further explore the practical impact of learning-based FOD enhancement in a real-world clinical study. Therefore, we hypothesized that an abnormal connection could serve as a potential biomarker, which should demonstrate a significant difference ($p < 0.05$) between patients and healthy controls in enhanced connectomes, similar to that seen with multi-shell-derived GTs. Furthermore, we also focused on a connection not expected to be affected, as this allows us to assess whether the enhancement process introduces potential false-positive (abnormal) connections, thereby ensuring specificity of the biomarker identification process.

Previous research studies have consistently highlighted deficits in the primary motor cortex among ALS patients~\cite{baumer2014advances}. Rose et al.~\cite{rose2012direct} proposed that measures of altered intra- and inter-hemispheric structural connectivity of the primary motor and somatosensory cortex would provide an improved assessment of corticomotor involvement in ALS. The findings demonstrate that the precentral gyrus interhemispheric connectivity would change between ALS patients and controls, while the postcentral gyrus interhemispheric connectivity would not change. 
To validate our hypothesis, we performed independent t-tests to measure whether specific structural connections from the primary motor cortex (i.e., left precentral cortex [L.PrCG] to the right precentral cortex [R.PrCG]) and from somatosensory cortex (i.e., left postcentral cortex [L.PoCG] to the right postcentral cortex [R.PoCG]) were significantly different between the ALS patient group and controls for each FOD method. Effect size was also computed to quantitatively measure the magnitude of the difference between groups. We also performed an analysis of sample size determination with respect to different levels of statistical power to assess whether FastFOD-Net could effectively facilitate clinical studies. Similarly, for the Tremor dataset, one-way ANOVA analysis was performed to determine whether there were significant differences in pathology-related connections, i.e., thalamus connectivity~\citep{kyle2023tremor}, between the means of three independent groups. 
\\

\noindent\textbf{Clinical correlation analysis}
For clinical validation in the MS cohort, we investigated the association between graph metrics derived from the estimated connectomes and conventional MS imaging markers (T2 lesion volume, whole-brain atrophy) as well as clinical metrics (Expanded Disability Status Scale [EDSS] and disease duration). Pearson’s correlation was used to assess these relationships, with statistical significance set at $p < 0.05$.

\bibliography{sn-article}

\clearpage
\section{Supplementary methods}

\subsection{Datasets}\label{sec:datasets}
The comparison between datasets used in FOD-Net~\citep{zeng2022fod} and our work is summarized in Supplementary Table~\ref{tab:datasets}.

\subsubsection{The Human Connectome Project}\label{sec:datasets_hcp} 

The Human Connectome Project (HCP)~\citep{van2013wu}, launched in 2010, stands out as one of the largest and most commonly used datasets within the neuroimaging community. For each subject, diffusion-weighted imaging (DWI) sequences were acquired using advanced multi-shell high angular resolution diffusion imaging (HARDI)protocols~\citep{sotiropoulos2013advances}. The HCP study aimed to study human brain connectivity from an initial cohort of 1,200 young adults aged between 22 and 35 years. One hundred unrelated samples were selected, and this dataset comprises minimally processed DWI data captured at an isotropic resolution of 1.25 mm with a 3 Tesla (3T) scanner. 
The repetition time (TR), echo time (TE), and field of view (FOV) were 5,520 ms, 89.5 ms and 256 mm, respectively. 
Three gradient shells with b-values of 1,000, 2,000, and 3,000 s/mm$^2$ were utilized, encompassing 90 diffusion-encoding directions per shell. The acquisition matrix was set to $145\times145$ to acquire 174 coronal slices for each subject. Furthermore, the protocol included 18 b0 volumes. 
Each subject also included a high-resolution T1-weighted dataset, which was acquired with an isotropic voxel size of 0.7 mm, TR/TE = 2,400/2.14 ms, and flip angle = $8\degree$.

Additionally, 15 HCP subjects with focal structural anomalies (denoted as HCP$_{Anomaly}$), but considered benign during structural quality control (QC)~\cite{van2013wu,power2017simple}, were assessed separately to evaluate the enhancement performance. The regions of interest (ROIs) were delineated by trained neuroimaging analysts and guided by the associated report\footnote{\url{https://www.humanconnectome.org/study/hcp-young-adult/document/1200-subjects-data-release}}. 

\subsubsection{Clinical datasets}\label{sec:datasets_clinical}
We also included three neurological study cohorts: (i) a tremor dataset (Tremor) encompassing patients who exhibited various tremor disorders, (ii) and a motor neuron disorder dataset (MND) including healthy controls and patients, and (iii) a multiple sclerosis dataset (MS) with clinical outcomes.

The Tremor dataset, described previously~\citep{kyle2023tremor}, contains patients with Parkinson's disease (PD), essential tremor (ET) and Dystonic tremor (DT). Imaging data were acquired from patients diagnosed with one of the neurological movement disorders at a single center using a 3T MRI scanner (SIGNA Architect, General Electric, Milwaukee). A total of 39 subjects with a consistent diffusion protocol were included, which consisted of 8 b0 volumes and 3 gradient shells of b-values of 700, 1,000, 2,800 s/mm$^2$ with 25, 40, and 75 directions, respectively, and the following acquisition parameters: TR = 6,250 ms, TE = 106 ms, flip angle = $90\degree$, FOV = 230 mm, acquisition matrix = $128 \times 128$, slice thickness = 2 mm, and a total of 72 axial slices. 
The structural imaging protocol consisted of a sagittal 3D IR-FSPGR T1-weighted image with the following parameters: TR = 8 ms, TE = 3.24 ms, flip angle = 10$\degree$, FOV = 256 mm, acquisition matrix = 256 $\times$ 256, slice thickness = 1.2 mm, and a total of 146 sagittal slices. The age range of this patient cohort spanned from 45 to 96 years, including a higher proportion of older individuals.

The second clinical dataset, denoted as MND, was collected from 97 samples across three different clinical groups using a 3T MRI scanner (GE MR750, DV29, 32-channel Nova head coil): 39 amyotrophic lateral sclerosis (ALS) subjects, 20 healthy controls (HCs), and 38 subjects with other motor neuron disorders. The age range of this cohort is from 42 to 79 years. The diffusion acquisition protocol includes four b-values of 0, 700, 1,000, 2,800 s/mm$^2$ with 8, 25, 40, and 75 directions, respectively. The parameters were set as follows: TE/TR = 100/3,245 ms, flip angle = 90$\degree$, FOV = 240 mm, acquisition matrix = 128 $\times$ 128, an isotropic voxel size of 2 mm, and a total of 70 axial slices. 
The T1-weighted images were acquired with a matrix size of 256 $\times$ 256 and an isotropic voxel size of 1 mm along the coronal orientation. Other parameters were set as follows: TE/TR = 2.3/6.2 ms, flip angle = $12\degree$, FOV = 240 mm, and the number of slices = 204.

The last MS dataset consists of 111 multiple sclerosis subjects with well-characterized clinical outcomes, such as lesion volume, brain atrophy, disease duration, and Expanded Disability Status Scale (EDSS) within 12 months. This population presented a large age range, from 17 to 78 years old. The dataset was acquired using a 3T Siemens Skyra scanner. The acquisition parameters of DWI were: TE/TR = 111/3,300 ms, flip angle = 90$\degree$, FOV = 256 mm, acquisition matrix = 128 $\times$ 128, slice thickness = 2 mm, number of slices = 76. Due to various differences in the diffusion acquisition protocol, we focused on cases that had 30 directions for the 1,000 and 2,500 b-values. The T1-weighted images were acquired with the matrix size of 256 $\times$ 256 and an isotropic voxel size of 1 mm along the sagittal orientation. Other parameters were set to an isotropic voxel size of 1 mm, TE/TR = 2.98/2,300 ms, flip angle = 9$\degree$, FOV = 256 mm, and the number of slices = 176.

\subsection{Data processing}\label{sec:data_processing}
The data processing workflow is outlined in Supplementary Fig.~\ref{fig:dwi_processing}. Since the DWI data are subject to various imaging artifacts, particularly the inherent subject motion and physiological noise, a range of preprocessing steps is recommended to reduce the confounding effects inherent in DWI acquisitions~\citep{tax2022whats}. 
Except for minimally processed data from HCP and HCP$_{Anomaly}$, other clinical datasets were preprocessed following recommendations by MRtrix~\citep{tournier2012mrtrix,tournier2019mrtrix3} and FSL~\citep{jenkinson2012fsl}. 
Denoising was the first step to be applied, since noise amplifications lower the sensitivity of DWI data. Then, unringing was performed to remove Gibbs ringing artifacts using the local subvoxel-shifts method~\citep{kellner2016gibbs}. Motion and distortion correction were conducted to address the misalignment of slices and microstructural parameters~\citep{tax2022whats,pierpaoli2010artifacts}. Finally, bias field correction was performed as a preprocessing step to eliminate low-frequency intensity inhomogeneities across the image~\citep{tustison2010n4itk}. 

Additionally, the T1-weighted images were segmented using FreeSurfer~\citep{fischl2012freesurfer} according to the Desikan-Killiany (DK) atlas~\citep{desikan2006automated}, to generate a parcellation image consisting of 84 discrete nodes. The full names of brain nodes and their corresponding lobes in the DK atlas is presented in Supplementary Table~\ref{tab:dk_fullname}. Pure and partial white matter (WM) tissues were defined through the segmentation of WM and gray matter (GM) using hybrid surface and volume segmentation (HSVS)~\citep{smith2020hybrid}. This segmentation can be used to construct a five-tissue-type (5TT) mask for each subject, encompassing cortical GM, subcortical GM, WM, cerebrospinal fluid (CSF), and pathological tissue (which remains empty in healthy subjects). Subsequently, the 5TT and the parcellation images were registered to the processed DWI and used for structural connectome reconstruction. 

To emulate single-shell low angular resolution diffusion imaging (LARDI) data, we used the Kennard-Stone subsampling strategy~\citep{kennard1969computer} on the original multi-shell HARDI diffusion data. This strategy enabled the selection of 32 uniformly distributed directions from the lowest b-value shell (b = 1,000 or 700 s/mm²; 30 directions for MS data). Subsequently, these subsets were combined with the b0 data. The subsequent step involves the generation of fiber orientation distribution (FOD) maps with MRtrix~\citep{tournier2012mrtrix,tournier2019mrtrix3}. First, the `dhollander' algorithm~\citep{dhollander2016novel} was used to estimate WM, GM, and CSF response functions for each subject and compute the group average across all subjects of each dataset~\citep{dhollander2021fixel}. 
Note that the response functions for the subsampled DWI data were recomputed since the downsampling process can introduce slight changes in voxel values on the DWI data, and ensuring accurate response functions was crucial for the subsequent modeling process. Then, two distinct operations were performed to generate the corresponding FOD maps: single-shell three-tissue (SS3T) constrained spherical deconvolution (CSD)~\citep{dhollander2016unsupervised} on the single-shell sampled data, and multi-shell multi-tissue (MSMT)~\citep{jeurissen2014multi} CSD on the multi-shell HARDI data. 
A maximum harmonics order ($l_{max}$) equal to 8~\citep{tournier2004direct,zeng2022fod} was used, yielding a total of 45 spherical harmonic coefficients regardless of the number of gradient directions used during acquisition. 

Fiber Orientation Distributions (FODs) are continuous functions that can be discretized into a finite number of fiber directions, referred to as `fixels', within each voxel~\citep{smith2013sift}. White matter bundles were segmented using fixel data derived from MSMT FODs through TractSeg~\citep{wasserthal2018tractseg}.

A connectome~\citep{yeh2019connectomes} is a complete map of brain connections, which can be used to investigate global changes in brain connectivity and probe the connectivity of specific regions of interest (ROIs) with the rest of the brain. The procedure for constructing the human connectome~\citep{yeh2019connectomes} of each subject consists of four primary stages: (1) whole-brain tractography; (2) quantifying the weight of each streamline to ensure quantitative connectivity results~\citep{smith2015sift2}; (3) selecting a GM atlas; (4) constructing the connectome by assessing the degree of structural connectivity between pairs of nodes.
For the first step, we employed anatomically constrained whole-brain tractography on all subjects using MRtrix~\citep{tournier2012mrtrix,tournier2019mrtrix3}. Driven by second-order integration over fiber orientation distributions (iFOD2), the tractogram under the framework of anatomically constrained
tractography (ACT)~\citep{smith2012anatomically} was computed by taking a given FOD image as the input and providing the corresponding 5TT mask. Notable parameters included a step size of 0.5 $\times$ voxel size, a maximum curvature of 45 degrees per step, a length range from 5 to 250 mm, and a FOD amplitude threshold of 0.06. Seeding was randomized from the interface between GM and WM, generating 10 million streamlines. 
Following tractography, the Spherical Deconvolution Informed Filtering of Tractograms 2 (SIFT2) algorithm~\citep{smith2015sift2} was applied to assign weights to each streamline, enhancing the quantitative and biologically relevant properties of the tractogram~\citep{smith2015effects}. 
Subsequently, in the third step, the T1-weighted image of each subject was segmented into meaningful brain nodes using the DK atlas as mentioned above~\citep{desikan2006automated}. In the fourth phase, we constructed the connectome matrix for each FOD-based tractogram by summing the SIFT2 weights of all streamlines connecting each pair of nodes.

\subsection{Architecture and implementation details}\label{sec:architecture}
FOD-Net \cite{zeng2022fod} applies a combined architecture of convolutional neural networks (CNNs) and multilayer perceptrons (MLP) to enhance coefficients of a single FOD estimate voxel-by-voxel and for each spherical harmonics order separately based on single-shell LARDI data. This approach is time consuming due to voxel-wise learning and high computational complexity when compared to convolutions, and overlooks the inherent 3D attributes of imaging data and inter-order coefficient relationships. Here, we extend the FOD-Net design to enable the direct prediction of entire patches by adopting a fully convolutional backbone~\citep{cciccek20163d}. 
This not only learns better features of FODs by overcoming the above limitations but also conserves computational resources, allowing for greater practical applicability.
Due to the capacity of U-Net to effectively handle diverse yet small datasets and yield acceptable predictions even with limited annotated data, researchers have proposed an appropriate training framework without architectural changes that automatically adapts to a given dataset for semantic segmentation~\citep{isensee2021nnu}. 
Here, we optimized the training paradigm for learning-based FOD enhancement. Firstly, to accelerate the convergence of our model and reduce its sensitivity to outliers, coefficient-wise standardization following multi-tissue informed intensity normalization \cite{dhollander2021multi} was undertaken on FODs and all coefficients were brought to a consistent scale. Moreover, to allow for the consideration of information from neighboring FODs within the same patch and interactive learning between different orders, our model abandoned separate order learning, and FOD estimates were trained on a patch-wise basis regardless of the original data size; this also makes the solution more practically applicable.
Coefficient-wise rescaling was eventually applied to the combined prediction to adjust each coefficient back to its original scale.

Our model was trained with the mean squared error (MSE) loss and Adam optimizer, with the number of epochs set to 50. The learning rate was set to 0.01 with a batch size of 4.
The `U' shape architecture includes a contracting path that captures features of FODs through 3D convolutional and downsampling layers, as well as an expansive path that upsamples feature maps to generate high-resolution FOD patches. The inclusion of skip connections facilitates the transfer of fine-grained information between the encoder and decoder. It consists of 4 encoding blocks, each corresponding to a decoding block at the same scale. The number of channels doubles with each increasing level during encoding, and it decreases by half with each subsequent level during decoding. Convolutional blocks with stride 2 were applied for downsampling, and interpolation was used for upsampling. We cropped input patches of $64 \times 64 \times 64 \times 45$ with a sliding window of stride 32 and then fed them into the model for training. 

\subsection{FOD metrics}
Angular correlation coefficient, denoted as $\mathbf{r_{Angular}}$~\citep{anderson2005measurement}, is calculated as follows:
\begin{equation}
    \mathbf{r_{Angular}}(u,v) = \frac{\sum_{c=1}^{C} u_{c} {v_{c}}^\mathbf{T}}
    {\sqrt{\sum_{c=1}^{C}{u_{c}}^2}\sqrt{\sum_{c=1}^{C}{v_{c}}^2}}
\label{eq:r_Angular}
\end{equation} 
\noindent where $u$ and $v$ refer to the ground truth (GT) FODs and the FODs estimated from LARDI (with or without enhancement), respectively; $C$ denotes the total number of spherical harmonic coefficients, and $c$ refers to the $c^{th}$ spherical harmonic coefficient. The coefficient provides a quantitative measure of the agreement between two FOD estimates.

\subsection{Fixel metrics}\label{sec:fixel_metrics}
To further assess the performance of enhancement frameworks on regions with varying fiber complexity, we generated masks based on TractSeg masks. Particularly, three main anatomical ROIs were defined~\citep{zeng2022fod}: a single-fiber region (ROI 1) within a selection of the corpus callosum (CC) primarily composed of large and coherent fiber tracts; the ROI 2 region where the middle cerebellar peduncle (MCP) and corticospinal tract (CST) bundles intersect, both of which exhibit coherent orientations, resulting in the voxels containing only two fixels in this intersection (ROI 2); and the intersected region of the superior longitudinal fascicle (SLF), CST, and CC with more complex fiber configurations (i.e., three distinct fixels, ROI 3). 
Since the regions were defined based on the number of crossing fiber bundles, we removed voxels that contained more fixels than expected for a fairer analysis regarding fixel complexity. For example, ROI 3 should only have three-fixel voxels. 

Mean angular error, denoted as $\mathbf{\mu E_{Angular}}$~\citep{zeng2022fod}, is defined as follows: 
\begin{equation} \label{eq:mae}
\mathbf{\mu E_{Angular}} = \frac{1}{n} \sum_{i=0}^{n} 
\frac{\mathbf{f}^\mathbf{\vec{d}} \cdot  \mathbf{\hat{f}}^\mathbf{\vec{d}}}
{\|\mathbf{f}^\mathbf{\vec{d}}\|_{2}\ \|\mathbf{\hat{f}}^\mathbf{\vec{d}}\|_{2}}.
\end{equation}
Here, $\mathbf{f}^\mathbf{\vec{d}}$ is a unit vector that represents the direction of an estimated fixel in three-dimensional space (i.e., $x$, $y$, $z$ axes), $\mathbf{\hat{f}}^\mathbf{\vec{d}}$ is the unit vector of its GT, $\cdot$ is the dot
product, and $\|\ \|_{2}$ is the L2 norm. 
To compare fixels fairly, we selected a maximum threshold of 45 degrees when matching fixels. Fixels with angular errors higher than that are considered misaligned for the rest of the paper.
For matched fixels, peak error is computed as the absolute difference between the amplitude at the maximum peak of the estimated fixels $\mathbf{f}^{\mathbf{Peak}}$ and their corresponding GT ${\mathbf{\hat{f}}}^{\mathbf{Peak}}$. 
Moreover, fiber density (FD) error for matched fixels is computed in a similar manner as the absolute difference between the integral of a given fixel estimate $\mathbf{f}^{\mathbf{FD}}$ and its corresponding GT ${\mathbf{\hat{f}}}^{\mathbf{FD}}$ to estimate the density of fibers in a specific region. 
Misaligned fixels are considered as either extra ${\mathbf{f_{+}}}$ or missing ${\mathbf{{f}_{-}}}$ fixels in estimations. The whole value is then summed to compute $\mathbf{E_{Peak}}$ and $\mathbf{E_{FD}}$ as formulated in Equation~\ref{eq:pe_afde}. 

\begin{equation} \label{eq:pe_afde}
\mathbf{E}
= \sum_{i=0}^{n} 
\big|\mathbf{f - {\hat{f}}} \big| 
+ \sum_{i=0}^{n} \mathbf{f_{+}}
+ \sum_{i=0}^{n} \mathbf{{f}_{-}}
\end{equation}

\subsection{Connectome metrics}\label{sec:connectome_metrics}
A structural connectome, $\mathbf{M}$, serves as a graph matrix with nodes denoting gray matter regions and edges denoting the brain connections between them.

We defined the disparity matrix as an edge-wise disparity measurement to assess the overall difference between a LARDI estimate ($\mathbf{M_{Est}}$) and its corresponding GT matrix ($\mathbf{M_\mathbf{GT}}$) measured as the mean absolute error. Subsequently, we computed the subject-wise average of mean edge disparities to summarize the overall difference, denoted as the mean connectome disparity ($\mathbf{\mu Disparity}$). 
We also extended the disparity metrics with edge-wise paired t-tests with $p < 0.05$ to focus on whether the disparity was significantly different for each edge. 
Finally, we adjusted with false discovery rate (FDR) correction for multiple comparisons considering the number of edges and counted the percentage of significantly different edges. Moreover, since we used weighted connectivity matrices (cf. binary matrices), we applied $\mathbf{{Kendall's}\ \tau}$ coefficient~\citep{kendall1938new}, to assess the ranking agreement and similarity of the connection strengths between the estimated connectomes and GTs. 
While edge weights indicate connectivity magnitude, the absolute values may be less informative than the relative rankings, making rank-order correlation a more meaningful metric than direct numerical comparison.

In addition, to analyze the connectome enhancement quality from a network perspective, several graph theory metrics~\citep{zhang2005general,fornito2016fundamentals} were computed via the Brain Connectivity Toolbox (BCT)~\citep{rubinov2010complex} and compared between methods. 
The difference ratio (DR) of the graph metrics of the reference multi-shell connectome and other estimates,expressed in percentages, was calculated as:
\begin{equation}
\mathbf{DR} = \frac{\mathbf{M_{Est}} - \mathbf{M_{GT}}}{{\mathbf{M_{GT}}}} \times 100\%.
\end{equation}
The graph metrics analyzed include (see~\cite{fornito2016fundamentals,rubinov2010complex} for further details):
\begin{itemize}
    \item \textbf{Global efficiency}: Quantifies how efficiently information or signals can be exchanged between all pairs of nodes in a brain network, providing insights into the overall interconnectedness and communication effectiveness of a brain connectome. 
    \item \textbf{Transitivity}: The transitivity is the ratio of triangles to triplets in the network; an alternative to the clustering coefficient.
    \item \textbf{Density}: The fraction of present connections to possible connections; connection weights are ignored in calculations.
    \item \textbf{Assortativity}: A correlation coefficient between the degrees of all nodes on two opposite ends of a link; a positive assortativity coefficient indicates that nodes tend to link to other nodes with the same or similar degree.
    \item \textbf{Mean betweenness centrality}: The fraction of all shortest paths in the network that contain a given node; nodes with high values of betweenness centrality participate in a large number of shortest paths.
    \item \textbf{Mean modularity}: The optimal community structure is a subdivision of the network into non-overlapping groups of nodes in a way that maximizes the number of within-group edges and minimizes the number of between-group edges; modularity is a statistic that quantifies the degree to which the network may be subdivided into such clearly delineated groups.
    \item \textbf{Mean strength}: Node strength is a measure of the importance or influence of a node within a network, and is computed as the sum of the weights of all the edges connected to a brain node; the strength of the entire brain connectome is determined by averaging the strengths of all brain nodes. 
    \item \textbf{Mean network degree}: The number of links connected to the node; in directed networks, the in-degree is the number of inward links and the out-degree is the number of outward links;  connection weights are ignored in calculations.
    \item \textbf{Mean clustering coefficient}: The fraction of triangles around a node, and is equivalent to the fraction of the node’s neighbors that are neighbors of each other.
    \item \textbf{Mean local efficiency}: The global efficiency (see above) computed on node neighborhoods; related to the clustering coefficient.
\end{itemize}


\subsection{Fixel-based analysis}\label{sec:connectome_accuracy_metrics}
To perform fixel-based analysis (FBA), as illustrated in Supplementary Fig.~\ref{fig:fba_pipeline}, we utilized the publicly available MRtrix software package~\citep{tournier2019mrtrix3,tournier2012mrtrix}. 
To compare the FBA results obtained from two population templates, we initially registered the population templates of each competing FOD estimation method into the multi-shell space. Subsequently, we warped the subject FOD images into this new template space. This warping allowed us to perform the statistical analysis in a common space for the reoriented fixels, assessing statistical differences between the patient group and the control group in terms of FD. 
However, an additional fixel-matching step is necessary before making comparisons. For that step, we followed the methodology described in the Supplementary Methods~\ref{sec:fixel_metrics}.
Eventually, we determined whether the fixels showing significant differences between patients and controls in the multi-shell FODs (GT) could also be detected in our enhanced FOD estimates. To that end, some common detections are described below.
\begin{itemize}
    \item $\mathbf{TP}$ represents the number of significantly different fixels correctly identified;
    \item $\mathbf{FN}$ represents the number of significantly different fixels that were not identified as such;
    \item $\mathbf{FP}$ represents the number of fixels incorrectly identified as significantly different when they are not;
    \item $\mathbf{TN}$ represents the number of fixels correctly identified as not significantly different.
\end{itemize}

From these, the ${\mathbf{Sensitivity}}$ metric was computed to measure the ability of a FOD enhancement technique to correctly identify fixels that were indeed significantly different. ${\mathbf{Specificity}}$ was also computed to measure the ability of each enhancement method to correctly identify fixels that were not significantly different. Finally, the ${\mathbf{Precision}}$ metric measures the reliability of detecting significantly different fixels using enhanced FODs (i.e., the ratio between true positives and all significantly different fixels) and evaluates how well the method avoids false positives. 
We also included the $\mathbf{F1}$ score as formulated. This metric is particularly valuable in scenarios where there is an imbalance between the two classes being classified, such as significantly different fixels and non-significantly different fixels, as it combines both ${\mathbf{Specificity}}$ and ${\mathbf{Sensitivity}}$.


\clearpage

\section{Supplementary Figures}

\begin{figure*}[h]
    \includegraphics[width=\textwidth]{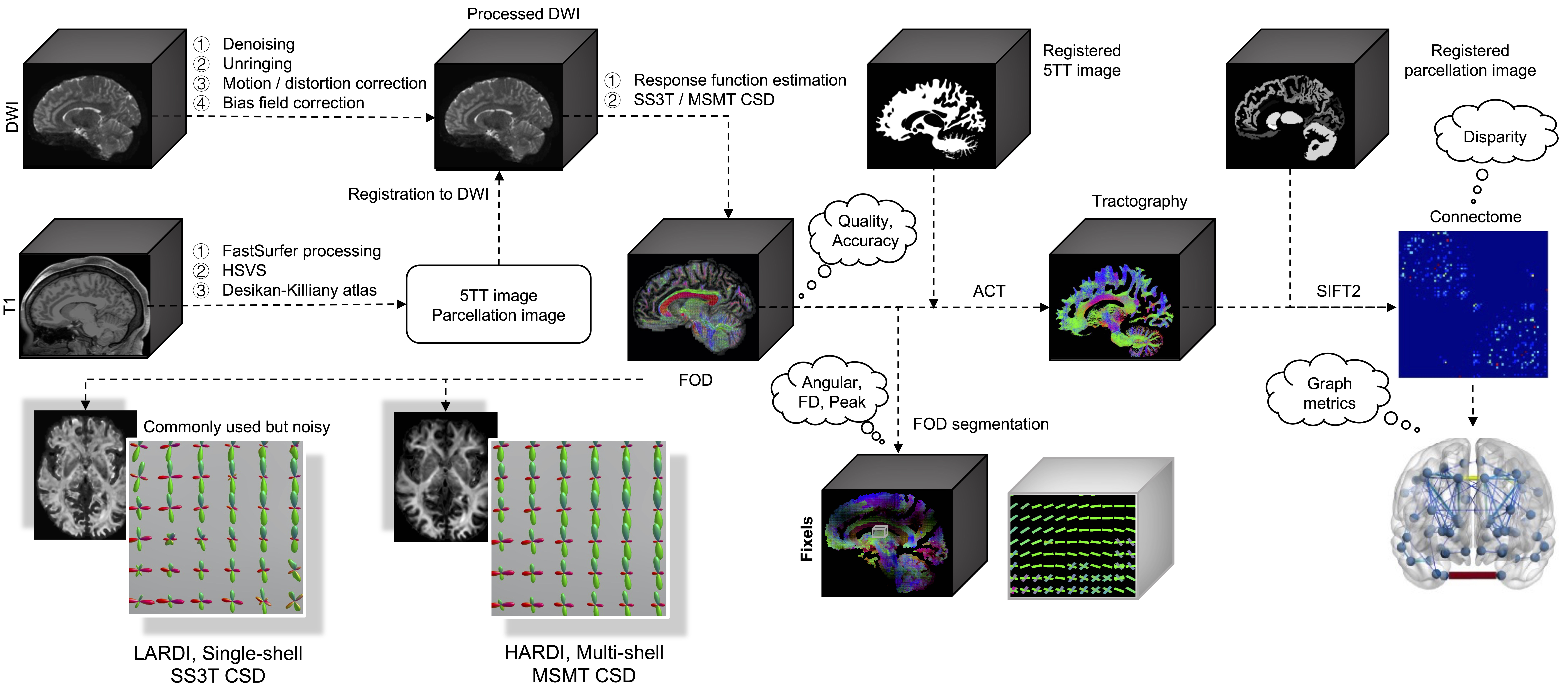}
    \caption{\textbf{Workflow of data processing.} The pipeline includes DWI preprocessing, T1-weighted image preprocessing. Fiber orientation distributions (FODs) are estimated using single-shell three-tissue (SS3T) or multi-shell multi-tissue (MSMT) constrained spherical deconvolution (CSD). FOD segmentation enables fiber bundle element `fixel' analysis from direction, fiber density (FD), and peak properties. Anatomicallyconstrained tractography (ACT) is performed utilizing registered five-tissue-type (5TT) images to generate tractograms. Fixel analysis can serve as a proxy evaluation of tractography. Finally, structural connectomes are constructed using registered parcellations and Spherical deconvolution Informed Filtering of Tractograms 2 (SIFT2) re-weighted streamlines, followed by analyses focusing on connectivity disparities and graph metrics.} 
\label{fig:dwi_processing}
\end{figure*}
\clearpage

\begin{figure*}[h]
    \centering
    \includegraphics[width=\linewidth]{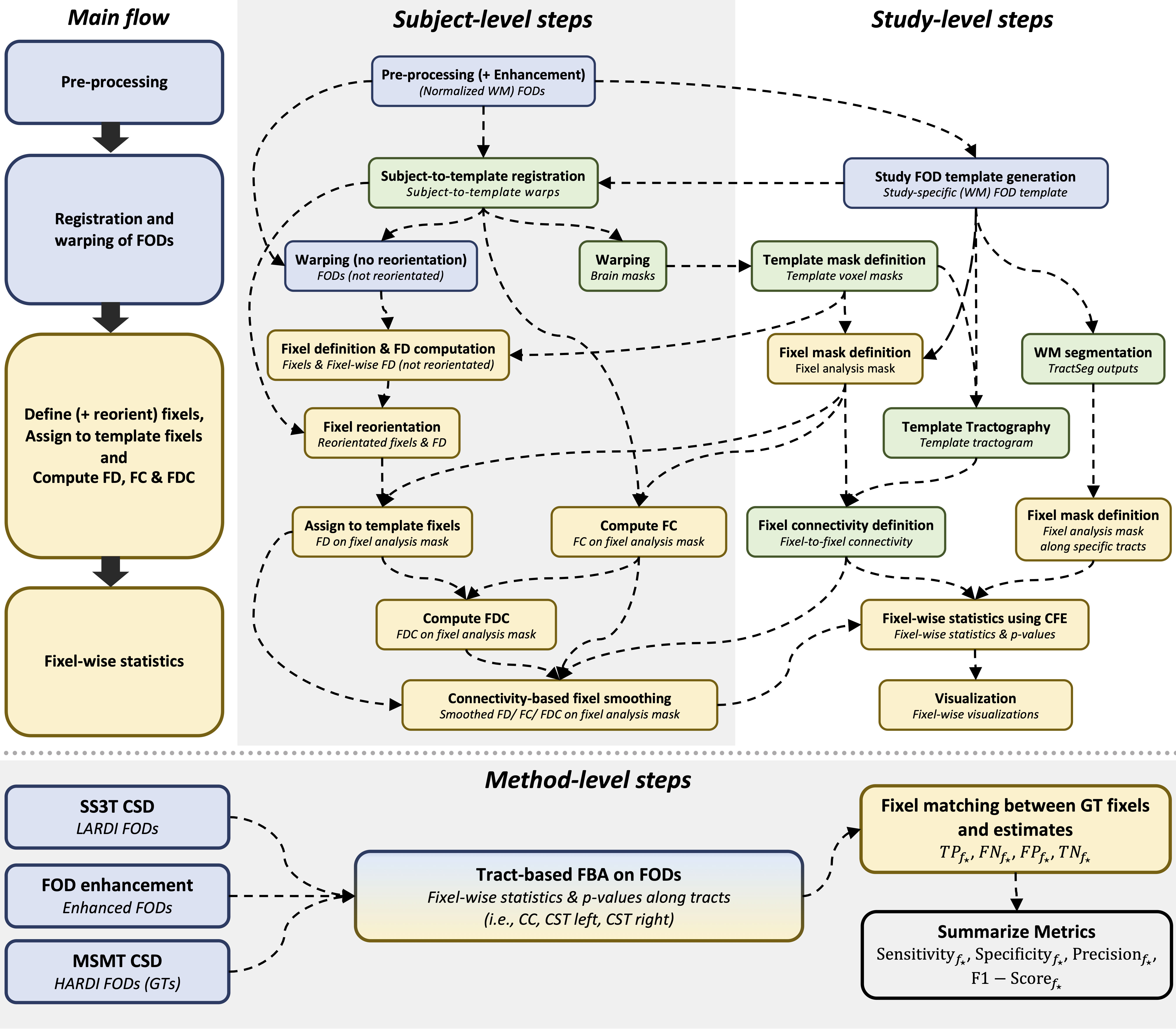}
    \caption{\textbf{Our fixel-based analysis (FBA) pipeline}.
 Blue boxes denote FOD data, and yellow boxes denote fixel-wise data. Each box names a processing step and its resulting output (\textit{italic}). Subject-level steps are performed once per individual subject, whereas study-level steps are operated only once for the study. For each FOD estimation method, i.e., SS3T CSD, FOD-Net, FastFOD-Net, and MSMT CSD, we performed a tract-based FBA study along specific tracts to generate fixel-wise statistics ($p < 0.05$) between clinical groups, and other subsequent method-level steps to compute quantitative metrics for comparisons between methods.} 
\label{fig:fba_pipeline}
\end{figure*}
\clearpage

\begin{figure*}[h]
        \includegraphics[width=\linewidth]{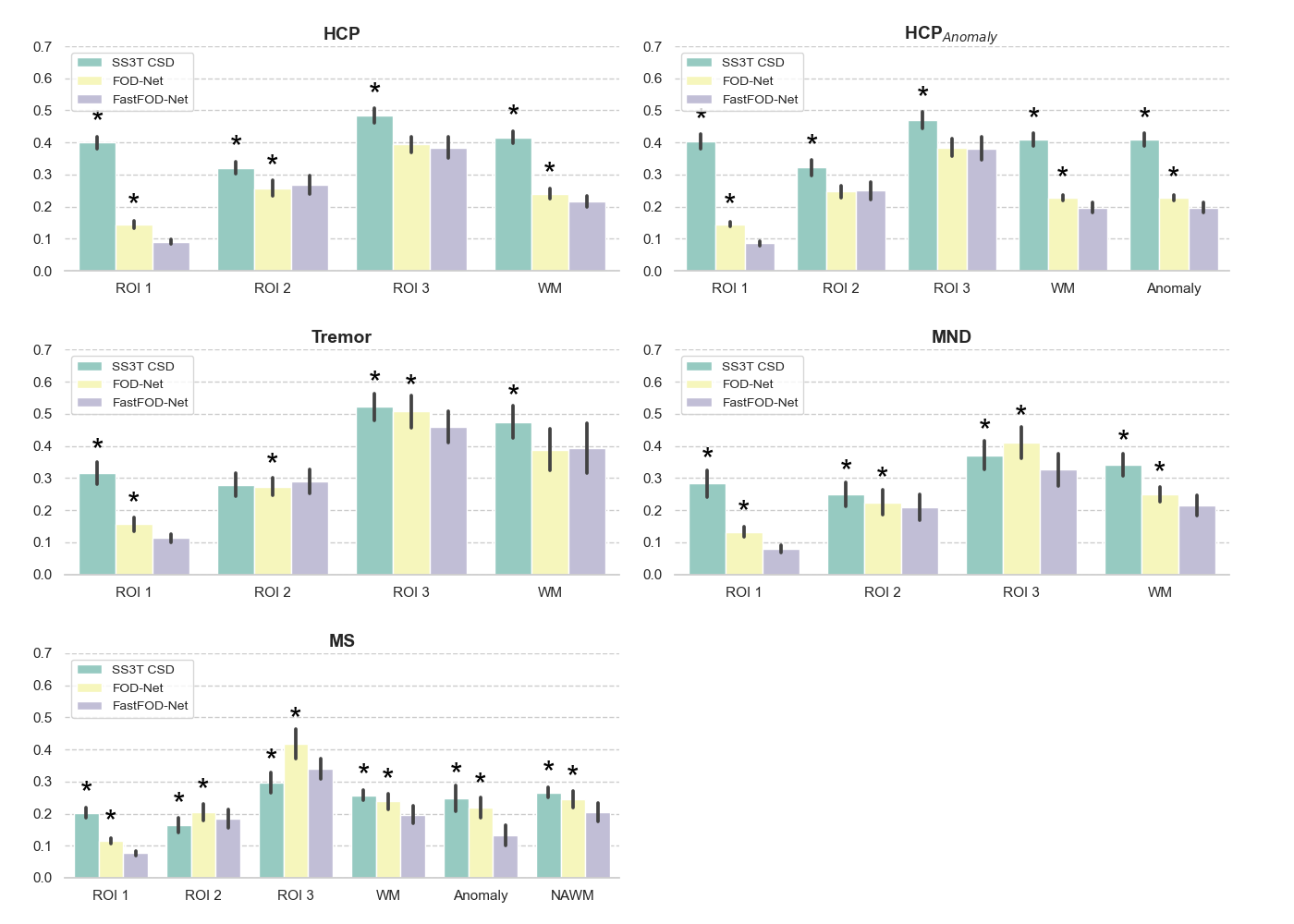}
        \caption{\textbf{Quantitative analysis of fixel assessment for the peak property.} Bar plots of peak error ($\mathbf{E_{Peak}}$) between ground truth (GT, corresponding to MSMT CSD) and other methods (SS3T CSD, FOD-Net, and FastFOD-Net) across patients and controls. ROI 1, 2, and 3 represent areas with distinct fiber configurations: single fiber population, two-way fiber crossing, and three-way fiber crossing, respectively. ‘Anomaly’ represents ROIs surrounding the anomalous regions in HCP$_{Anomaly}$ cases, and lesions in MS cases. WM denotes white matter, and NAWM represents normal-appearing white matter outside lesions. } 
\label{fig:peak_error}
\end{figure*}
\clearpage

\begin{figure*}[h]
    \includegraphics[width=\linewidth]{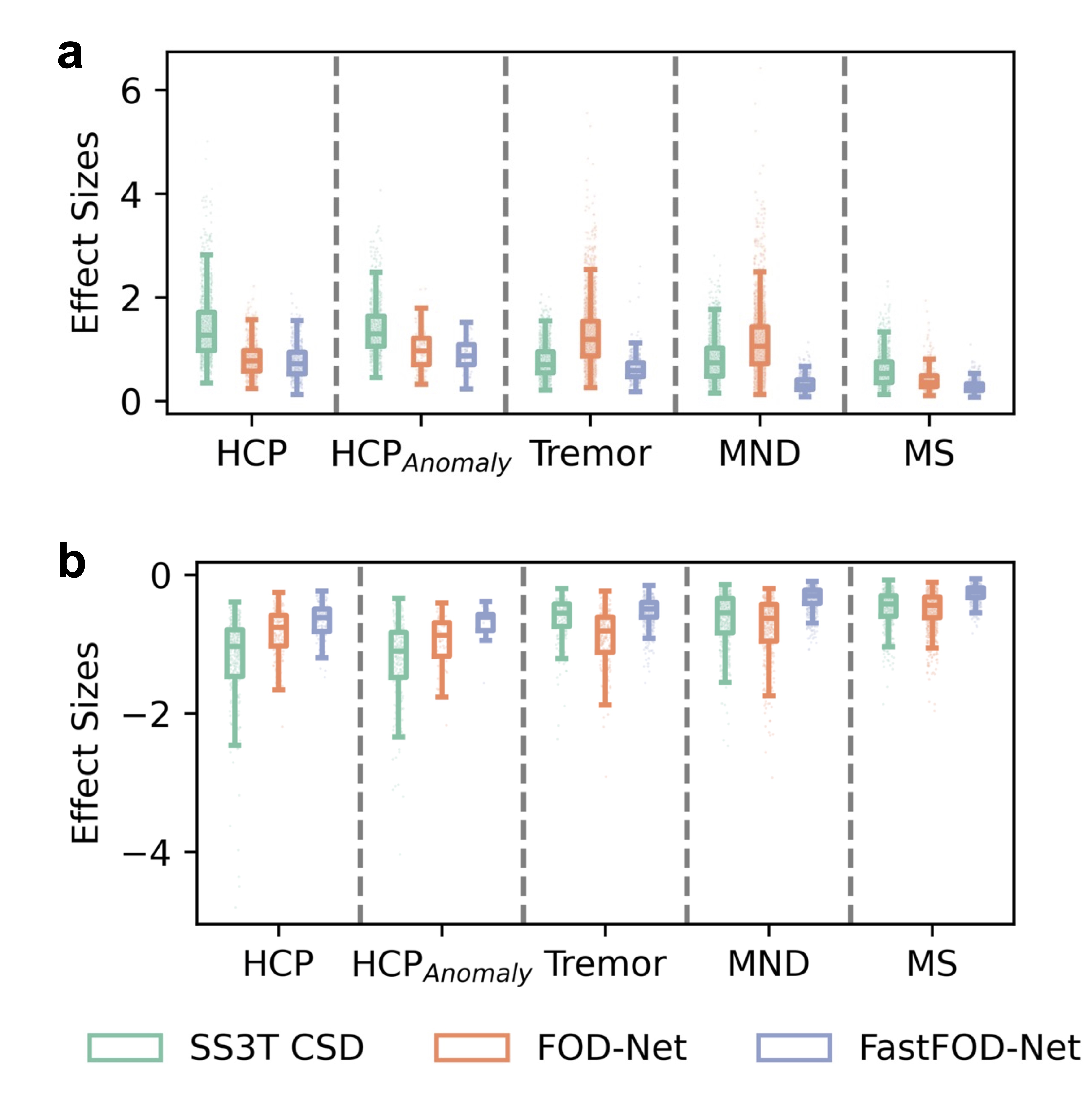}
    \caption{\textbf{Effect sizes of significantly different edges in connectomes between MSMT CSD (ground truth, GT) and alternative methods (SS3T CSD, FOD-Net, and FastFOD-Net).} \textbf{a,} Positive effect sizes indicate MSMT CSD yields higher connectivity values compared to other methods. \textbf{b,} Negative effect sizes indicate MSMT-CSD yields lower connectivity values compared to other methods. Box plots represent the distribution of effect sizes across different datasets.}
\label{fig:supp_significantedges}
\end{figure*}
\clearpage

\begin{figure*}[h]
    \includegraphics[width=\linewidth]{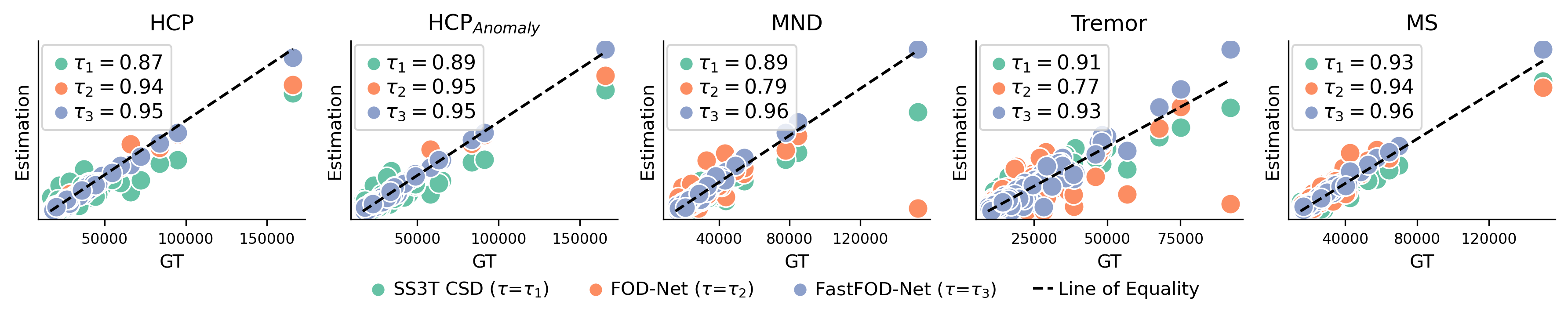}
    \caption{\textbf{Correlation analysis of mean connectome matrices between ground truth (GT) via MSMT CSD and other estimates via SS3T CSD, FOD-Net, and FastFOD-Net on HCP, HCP$_{Anomaly}$, Tremor, MND, and MS cases.} $\tau$ denotes Kendall ranking correlation coefficient.}
\label{fig:supp_connectome_estvsgt}
\end{figure*}
\clearpage

\begin{figure*}[h]
   \includegraphics[width=\linewidth]{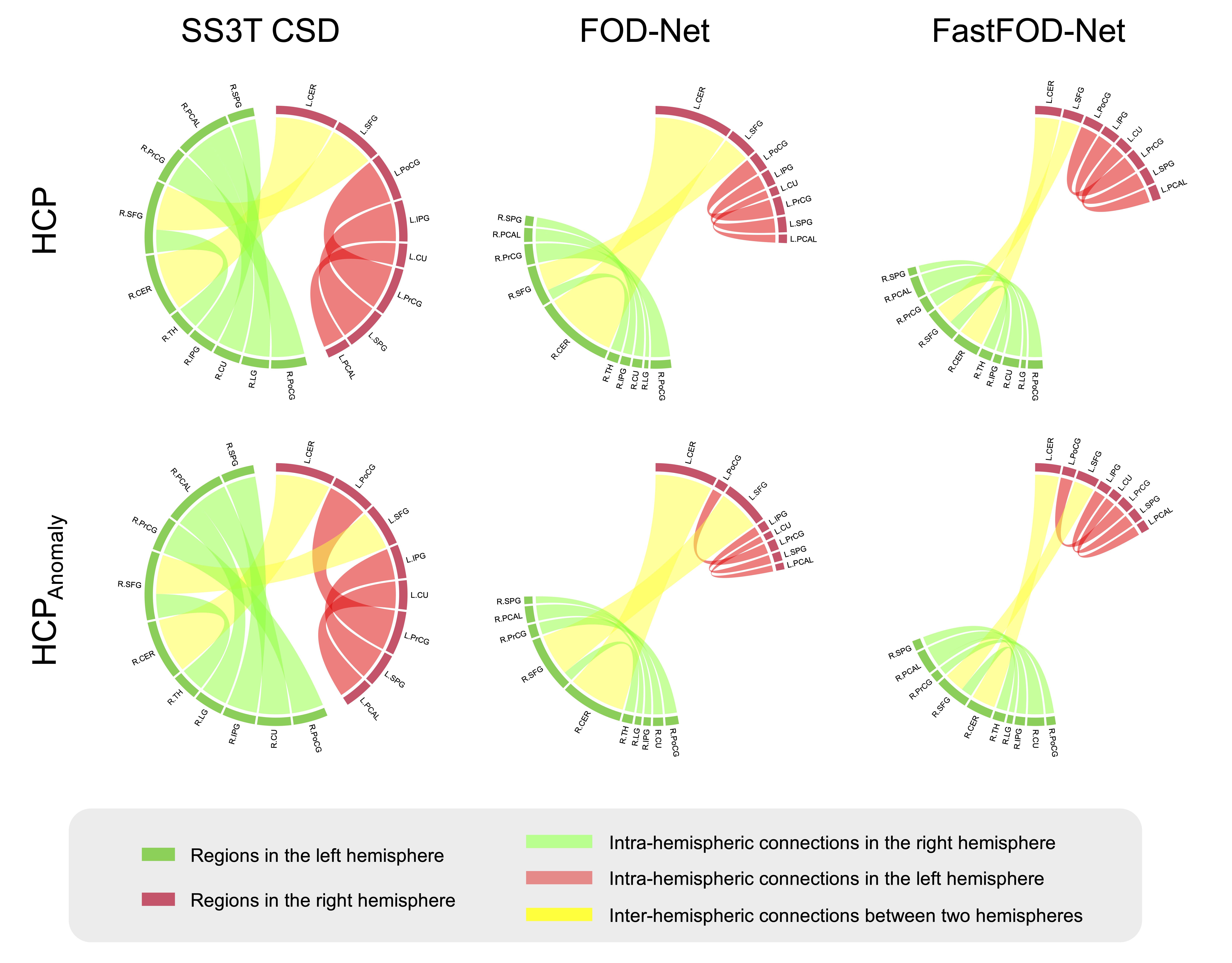}
    \caption{\textbf{Chord diagrams illustrating the top 10 connections with the largest disparities between ground truth (GT, MSMT CSD) and other methods (SS3T CSD, FOD-Net, and FastFOD-Net) on both the HCP and HCP$_{Anomaly}$ datasets.} 
    These connections were selected based on the edge disparity matrices comparing SS3T CSD to the GT. Nodes represent brain regions, with intra-hemispheric connections within the left hemisphere depicted in green, those within the right hemisphere in red, and inter-hemispheric connections in yellow. For full names of nodes, refer to Supplementary Table~\ref{tab:dk_fullname}.} 
\label{fig:supp_chord_disp_hcp}
\end{figure*}
\clearpage

\begin{figure*}[h]
        \includegraphics[width=\linewidth]{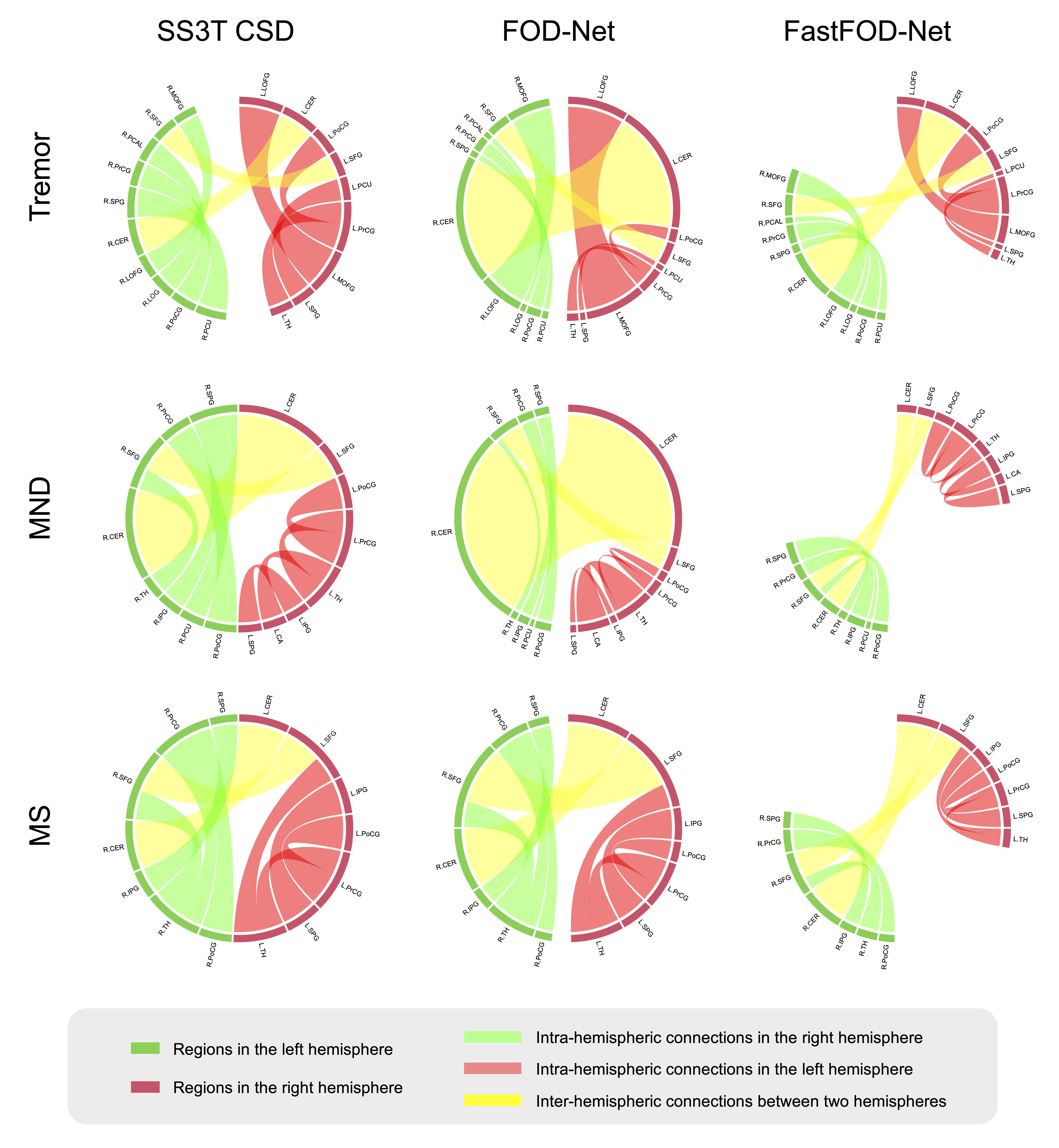}
        \caption{\textbf{Chord diagrams illustrating the top 10 connections with the largest disparities between ground truth (GT, MSMT CSD) and other methods (SS3T CSD, FOD-Net, and FastFOD-Net) on Tremor, MND, and MS dataset.}
        These connections were selected based on the edge disparity matrices comparing SS3T CSD to the GT. Nodes represent brain regions, with intra-hemispheric connections within the left hemisphere depicted in green, those within the right hemisphere in red, and inter-hemispheric connections in yellow. For full names of nodes, refer to Supplementary Table~\ref{tab:dk_fullname}.} 
\label{fig:supp_chord_disp_patient}
\end{figure*}
\clearpage



\begin{figure*}[t]
    \includegraphics[width=\linewidth]{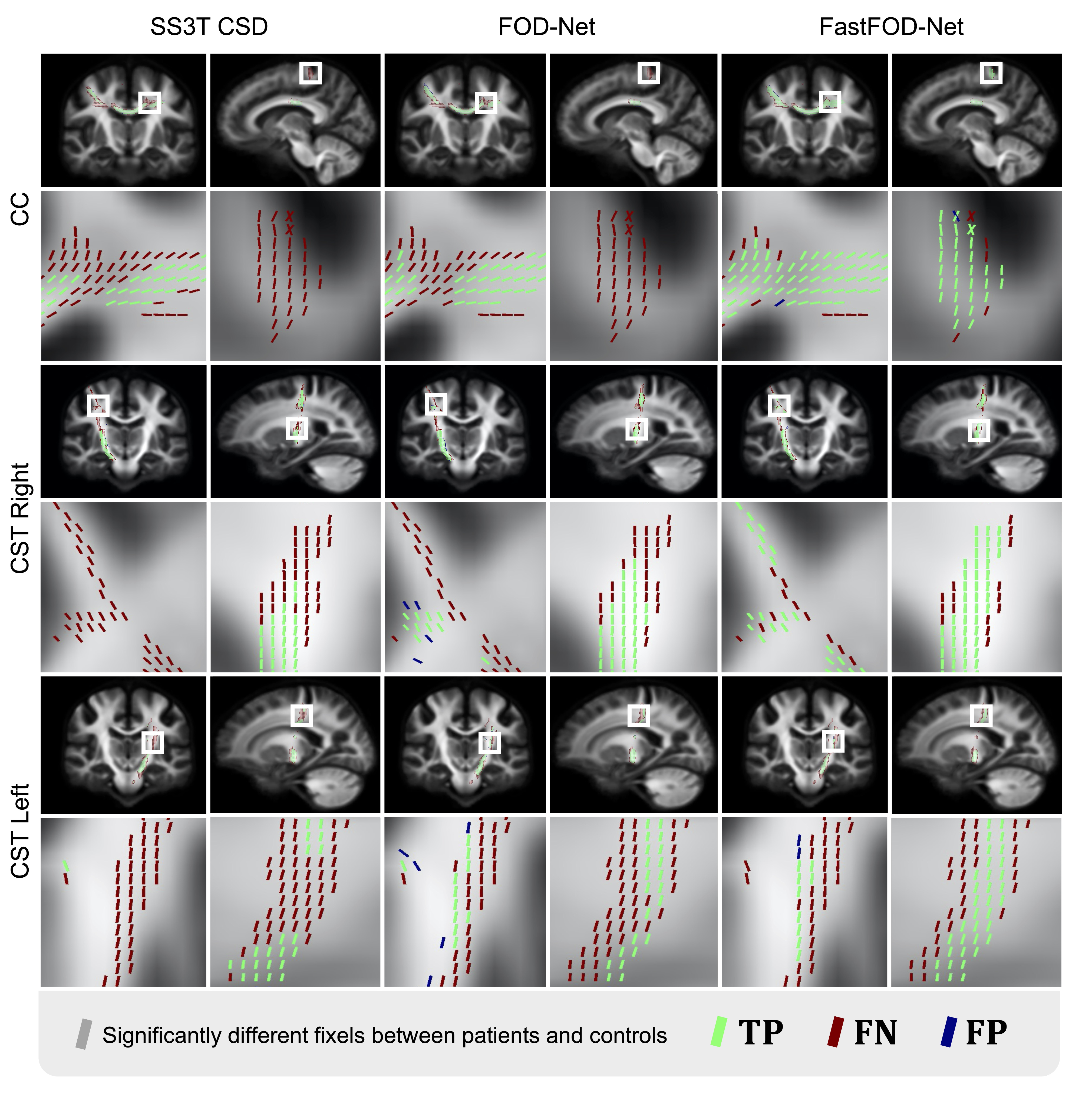}
    \caption{\textbf{Comparisons of significantly different fixels between the fiber density (FD) of amyotrophic lateral sclerosis (ALS) patients and controls using MSMT CSD as the ground truth (GT) method.} In this comparison, significant fixels derived from the GT method were juxtaposed with those obtained from alternative methodologies (specifically, SS3T CSD, FOD-Net, and FastFOD-Net).
    Notably, fixels correctly identified ($\mathbf{TP}$) by the comparison methods along various white matter tracts, namely the corpus callosum (CC), right corticospinal tract (CST), and left CST, were denoted by green coloring. Conversely, fixels that were not identified despite being significantly different with MSMT CSD ($\mathbf{FN}$) were depicted in red, while those erroneously identified as significant ($\mathbf{FP}$) were marked in blue. True negative fixels ($\mathbf{TN}$) are omitted for clarity, focusing on detected and real differences.}
    \label{fig:supp_fbavis}
\end{figure*}


\begin{figure*}[t]
    \centering
    \includegraphics[width=\linewidth]{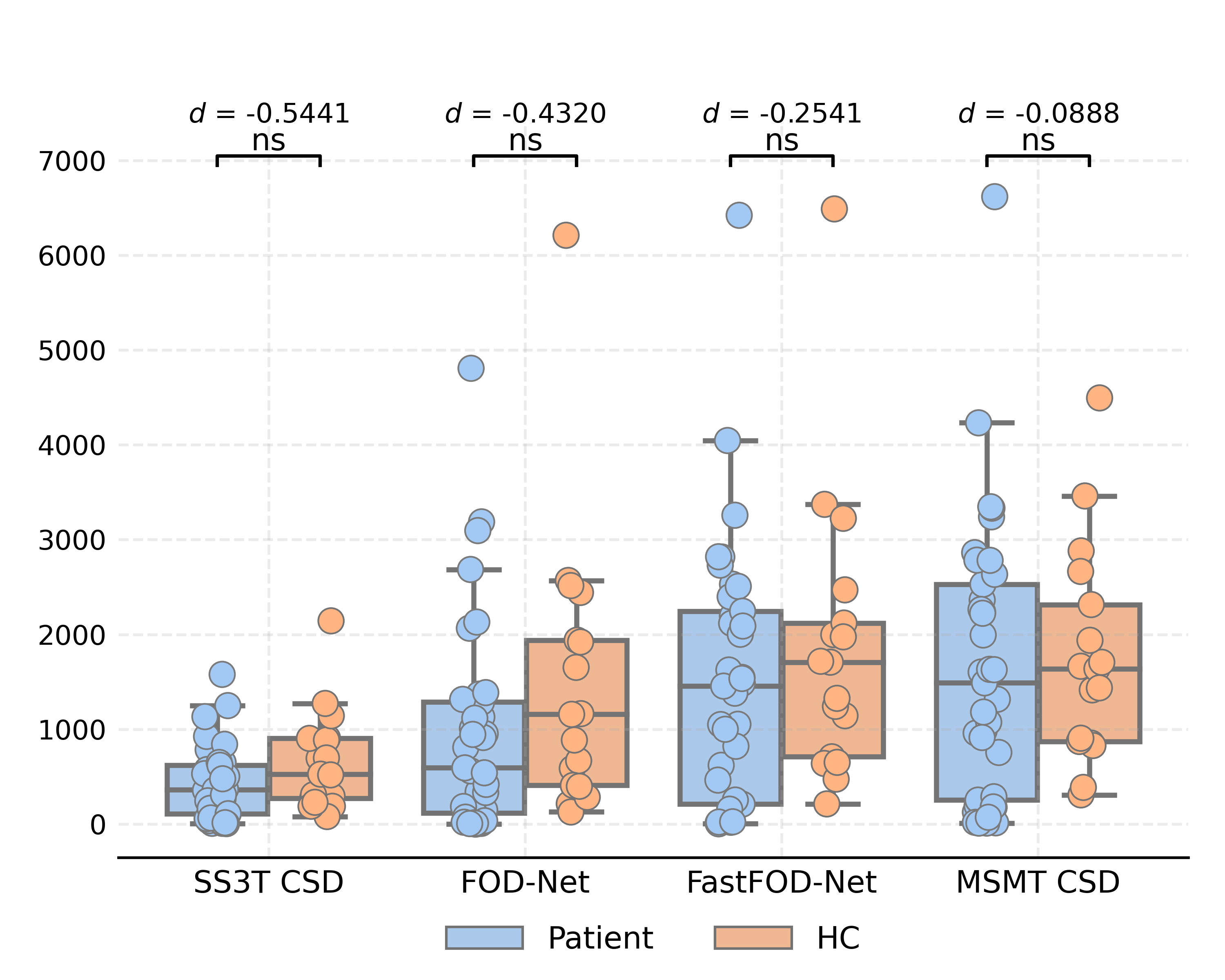}
    \caption{\textbf{Box plots depicting connectivity values  between amyotrophic lateral sclerosis (ALS) subjects and healthy controls (HCs) for a connection with no clinical significance: from left postcentral cortex (L.PoCG) to right postcentral cortex (R.PoCG)~\cite{rose2012direct}.} ALS samples are represented in blue, while HCs are shown in orange. Above each pair of box plots, p-values from independent t-tests indicate the statistical significance of group differences ($p < 0.05$ denoted by `*', while `ns' indicates non-significance). Cohen’s $d$ effect sizes further quantify the magnitude of these differences.} 
    \label{fig:connectome_pvshc_controlconnection}
\end{figure*}

\begin{figure*}[t]
\centering
    \includegraphics[width=\linewidth]{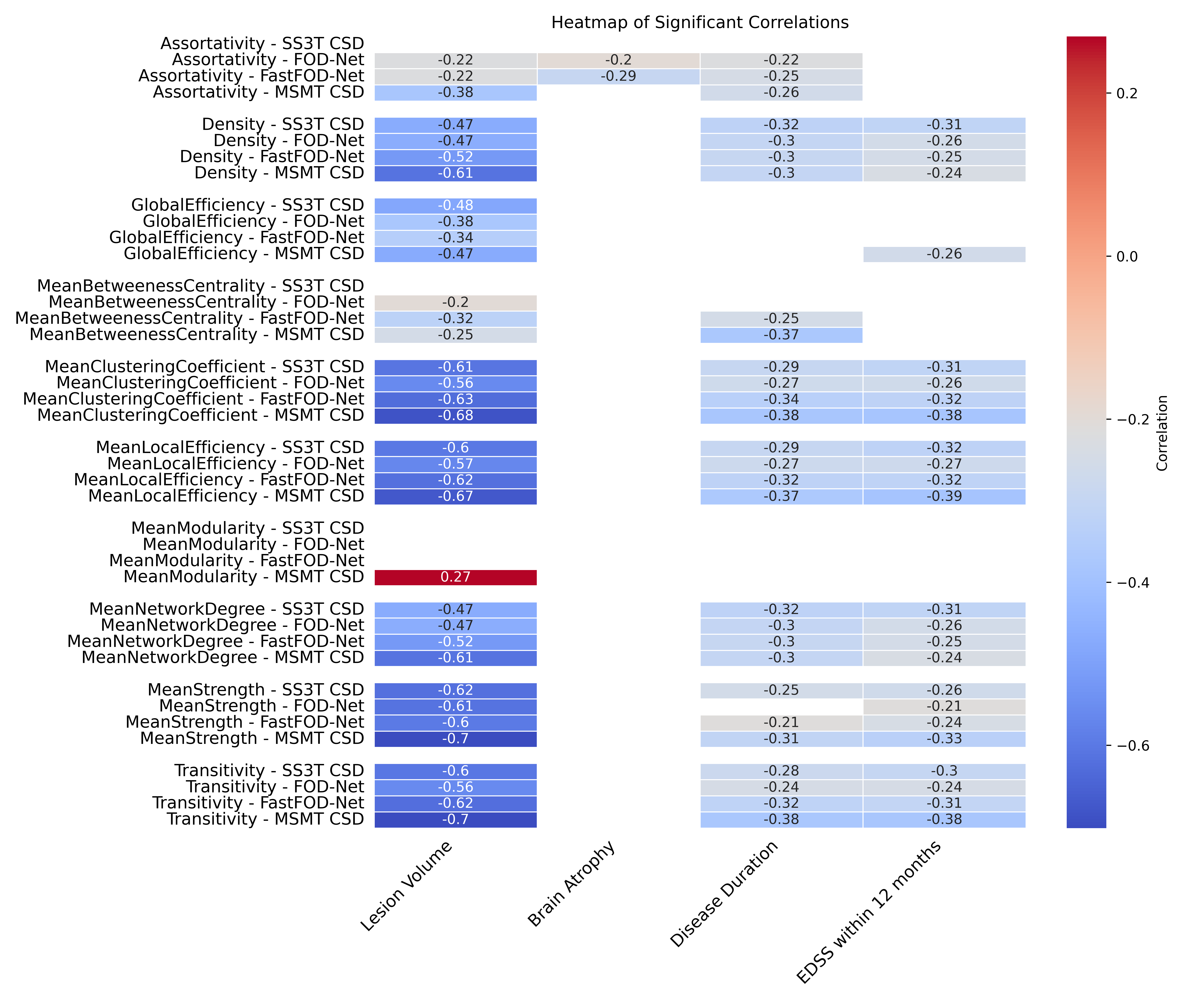}
    \caption{\textbf{Heatmap of significant correlations ($p < 0.05$) between graph metrics from different FOD estimation methods and clinical outcomes in multiple sclerosis patients.} Correlations are shown for total lesion volume, whole-brain atrophy, disease duration, and Expanded Disability Status Scale (EDSS) within 12 months.} 
\label{fig:graphmetics_with_clinicaloutcome}
\end{figure*}

\thispagestyle{empty}

\clearpage
\section{Supplementary Tables}

\begin{table*}[h]
		\caption{\textbf{Details of the datasets used in FOD-Net~\citep{zeng2022fod} and our work, including the acquisition parameters all pertaining to the diffusion imaging protocols.} HC denotes healthy control, while HC$_{Anomaly}$ represents healthy cases with focal brain abnormalities. For the Tremor dataset, DT, ET, and PD denote patient cases with Parkinson’s disease (PD), essential tremor (ET) and dystonic tremor (DT). For the MND dataset, ALS denotes amyotrophic lateral sclerosis cases, while `other' refers to other motor neuron disorder patients. The MS dataset consists of patients diagnosed with multiple sclerosis.}\label{dataset}
		\centering
            \begin{adjustbox}{width=\textwidth}
		\begin{tabular}{{c|c c c c |c c c c c }}
			\toprule
			\multicolumn{1}{c}{}&\multicolumn{4}{c}{\textbf{FOD-Net}} & \multicolumn{5}{c}{\textbf{FastFOD-Net}} \\
			\hline
			\hline
                Name & HCP & HCP$_{Anomaly}$ & Local & Multicenter & HCP & HCP$_{Anomaly}$ & Tremor & MND & MS \\
                Type & HC & HC$_{Anomaly}$ & HC & HC & HC & HC$_{Anomaly}$ & DT, ET, PD & ALS, other, HC  & MS \\
			\#Data$^{1}$ &  110 & 3 & 4 & 3 & 100 & 15 & 39 & 97 & 111 \\
                Age 
                & \multicolumn{2}{c}{22$-$35} 
                & 25$-$35
                & 23, 26, 23 
                    & \multicolumn{2}{c}{22$-$35} & 45$-$96 & 42$-$79 & 17$-$78 \\
                Spacing$^{2}$ & \multicolumn{2}{c}{1.25 mm} & 2 mm & - 
                    & \multicolumn{2}{c}{1.25 mm} & 1.8 mm & 2 mm &  2 mm\\
                Matrix$^{3}$ & \multicolumn{2}{c}{$145\times145$} & $128\times128$ & - 
                    & \multicolumn{2}{c}{$145\times145$} & $128\times128$ & $128\times128$ & $128\times128$ \\
                \#Slices$^{4}$ & \multicolumn{2}{c}{174} & 70 & - 
                    & \multicolumn{2}{c}{174} & 72 & 70 & 76 \\
                Orientation & \multicolumn{2}{c}{Coronal} & Axial & Axial 
                    & \multicolumn{2}{c}{Coronal} & Axial & Axial & Axial \\
                Shell bvalues$^{5}$ & \multicolumn{3}{c}{0, 1000, 2000, 3000} & - & \multicolumn{2}{c}{0, 1000, 2000, 3000} & \multicolumn{2}{c}{0, 700, 1000, 2800} & 0, 200, 500, 1000, 2500 \\
                Shell sizes$^{6}$ & \multicolumn{2}{c}{18, 90, 90, 90} & 8, 32, 32, 72 & - 
                    & \multicolumn{2}{c}{18, 90, 90, 90} & \multicolumn{2}{c}{8, 25, 40, 75} & 5/11, 3, 6 30, 30  \\
                Clinical outcome & - & - & - & - & - & - & - & - & $\checkmark$ \\
			\bottomrule
		\end{tabular}
            \end{adjustbox}
		\footnotesize
		\begin{tablenotes}
		\setlength{\itemindent}{0.in}
		    \item $^{1}$ \#Data represents the number of samples in this dataset. $^{2}$ Spacing is the voxel size. 
                \item $^{3}$ Size and $^{4}$ \#Slices denote the size of the acquisition matrix and the number of slices acquired along the orientation. 
                \item $^{5}$ Shell b-values list the average b-value for each shell, and $^{6}$ shell sizes list the number of volumes in each shell (from lowest to highest). 
		\end{tablenotes}
\label{tab:datasets}
\end{table*}

\clearpage
\begin{table*}[h]
    \caption{\textbf{Full names of brain regions and their corresponding lobes in the Desikan-Killiany (DK) atlas~\cite{desikan2006automated}.}}
    \centering
    \begin{adjustbox}{width=\textwidth}
    \begin{tabular}{llllll}
        \textbf{Key} & \textbf{Name}                        & \textbf{Short Name} & \textbf{Key} & \textbf{Name}                         & \textbf{Short Name} \\
        0            & Left-BanksSTS-Cortex                 & L.BSTS              & 42           & Right-Thalamus                        & R.TH                \\
        1            & Left-CaudalAnteriorCingulate-Cortex  & L.CACG              & 43           & Right-Caudate                         & R.CA                \\
        2            & Left-CaudalMiddleFrontal-Cortex      & L.CMFG              & 44           & Right-Putamen                         & R.PU                \\
        3            & Left-Cuneus-Cortex                   & L.CU                & 45           & Right-Pallidum                        & R.PA                \\
        4            & Left-Entorhinal-Cortex               & L.EC                & 46           & Right-Hippocampus                     & R.HI                \\
        5            & Left-Fusiform-Cortex                 & L.FG                & 47           & Right-Amygdala                        & R.AM                \\
        6            & Left-InferiorParietal-Cortex         & L.IPG               & 48           & Right-Accumbens                       & R.AC                \\
        7            & Left-InferiorTemporal-Cortex         & L.ITG               & 49           & Right-BanksSTS-Cortex                 & R.BSTS              \\
        8            & Left-IsthmusCingulate-Cortex         & L.ICG               & 50           & Right-CaudalAnteriorCingulate-Cortex  & R.CACG              \\
        9            & Left-LateralOccipital-Cortex         & L.LOG               & 51           & Right-CaudalMiddleFrontal-Cortex      & R.CMFG              \\
        10           & Left-LateralOrbitoFrontal-Cortex     & L.LOFG              & 52           & Right-Cuneus-Cortex                   & R.CU                \\
        11           & Left-Lingual-Cortex                  & L.LG                & 53           & Right-Entorhinal-Cortex               & R.EC                \\
        12           & Left-MedialOrbitoFrontal-Cortex      & L.MOFG              & 54           & Right-Fusiform-Cortex                 & R.FG                \\
        13           & Left-MiddleTemporal-Cortex           & L.MTG               & 55           & Right-InferiorParietal-Cortex         & R.IPG               \\
        14           & Left-ParaHippocampal-Cortex          & L.PHIG              & 56           & Right-InferiorTemporal-Cortex         & R.ITG               \\
        15           & Left-ParaCentral-Cortex              & L.PaCG              & 57           & Right-IsthmusCingulate-Cortex         & R.ICG               \\
        16           & Left-ParsOpercularis-Cortex          & L.POP               & 58           & Right-LateralOccipital-Cortex         & R.LOG               \\
        17           & Left-ParsOrbitalis-Cortex            & L.POR               & 59           & Right-LateralOrbitoFrontal-Cortex     & R.LOFG              \\
        18           & Left-ParsTriangularis-Cortex         & L.PTR               & 60           & Right-Lingual-Cortex                  & R.LG                \\
        19           & Left-Pericalcarine-Cortex            & L.PCAL              & 61           & Right-MedialOrbitoFrontal-Cortex      & R.MOFG              \\
        20           & Left-PostCentral-Cortex              & L.PoCG              & 62           & Right-MiddleTemporal-Cortex           & R.MTG               \\
        21           & Left-PosteriorCingulate-Cortex       & L.PCG               & 63           & Right-ParaHippocampal-Cortex          & R.PHIG              \\
        22           & Left-PreCentral-Cortex               & L.PrCG              & 64           & Right-ParaCentral-Cortex              & R.PaCG              \\
        23           & Left-PreCuneus-Cortex                & L.PCU               & 65           & Right-ParsOpercularis-Cortex          & R.POP               \\
        24           & Left-RostralAnteriorCingulate-Cortex & L.RACG              & 66           & Right-ParsOrbitalis-Cortex            & R.POR               \\
        25           & Left-RostralMiddleFrontal-Cortex     & L.RMFG              & 67           & Right-ParsTriangularis-Cortex         & R.PTR               \\
        26           & Left-SuperiorFrontal-Cortex          & L.SFG               & 68           & Right-Pericalcarine-Cortex            & R.PCAL              \\
        27           & Left-SuperiorParietal-Cortex         & L.SPG               & 69           & Right-PostCentral-Cortex              & R.PoCG              \\
        28           & Left-SuperiorTemporal-Cortex         & L.STG               & 70           & Right-PosteriorCingulate-Cortex       & R.PCG               \\
        29           & Left-SupraMarginal-Cortex            & L.SMG               & 71           & Right-PreCentral-Cortex               & R.PrCG              \\
        30           & Left-FrontalPole-Cortex              & L.FP                & 72           & Right-PreCuneus-Cortex                & R.PCU               \\
        31           & Left-TemporalPole-Cortex             & L.TP                & 73           & Right-RostralAnteriorCingulate-Cortex & R.RACG              \\
        32           & Left-TransverseTemporal-Cortex       & L.TTG               & 74           & Right-RostralMiddleFrontal-Cortex     & R.RMFG              \\
        33           & Left-Insula-Cortex                   & L.IN                & 75           & Right-SuperiorFrontal-Cortex          & R.SFG               \\
        34           & Left-Cerebellum-Cortex               & L.CER               & 76           & Right-SuperiorParietal-Cortex         & R.SPG               \\
        35           & Left-Thalamus                        & L.TH                & 77           & Right-SuperiorTemporal-Cortex         & R.STG               \\
        36           & Left-Caudate                         & L.CA                & 78           & Right-SupraMarginal-Cortex            & R.SMG               \\
        37           & Left-Putamen                         & L.PU                & 79           & Right-FrontalPole-Cortex              & R.FP                \\
        38           & Left-Pallidum                        & L.PA                & 80           & Right-TemporalPole-Cortex             & R.TP                \\
        39           & Left-Hippocampus                     & L.HI                & 81           & Right-TransverseTemporal-Cortex       & R.TTG               \\
        40           & Left-Amygdala                        & L.AM                & 82           & Right-Insula-Cortex                   & R.IN                \\
        41           & Left-Accumbens                       & L.AC                & 83           & Right-Cerebellum-Cortex               & R.CER              
    \end{tabular}
\end{adjustbox}
\label{tab:dk_fullname}
\end{table*}

\begin{table*}[t]
    \caption{\textbf{Quantitative analysis of mean peak signal-to-noise ratio$^{1}$ and angular correlation coefficient$^{2}$ of estimated FODs when compared to MSMT CSD-derived data$^{3}$.}} 
    \label{tab:fod_psnr_ra}
    \centering
    \setlength{\extrarowheight}{1.7pt}
    \begin{adjustbox}{width=\textwidth}
        \begin{threeparttable}
            \begin{tabular}{lllllll}
            \toprule
            & & \multicolumn{5}{@{}c@{}}{$\mathbf{PSNR}$\footnotemark[1] ($\uparrow$)} \\\cmidrule{3-7}%
            Methods & Tissue Type & \multicolumn{1}{l}{HCP} & \multicolumn{1}{l}{HCP$_{Anomaly}$$^{4}$} & \multicolumn{1}{l}{Tremor} & \multicolumn{1}{l}{MND} & \multicolumn{1}{l}{MS} \\
            \midrule
SS3T CSD~\cite{dhollander2016unsupervised} 
& WM\footnotemark[5] & 70.7430* ± 0.6071 & 70.6472* ± 0.9050 & 66.3592* ± 1.0522 & 72.8055* ± 1.0879 & 72.3808* ± 0.7498  \\ 
& JCWM\footnotemark[6] & 60.2084* ± 0.5640 & 60.3150* ± 0.9830 & 58.8588* ± 0.7082 & 64.2163* ± 1.2248 & 62.6626* ± 0.8832  \\ 
& WM-CGM\footnotemark[7] & 55.2939* ± 0.7760 & 55.2862* ± 0.7453 & 52.7151* ± 1.4621 & 57.0784* ± 1.0483 & 56.4695* ± 0.8889  \\ 
& Anomaly & NA & 70.6472* ± 0.9050 & NA & NA & 50.6064* ± 5.4511  \\ 
& NAWM & NA & NA & NA & NA & 72.6303* ± 0.6927  \\ 
FOD-Net~\cite{zeng2022fod}
& WM & 74.3551* ± 0.6627 & 74.3509* ± 0.9020 & 68.4481* ± 1.0767 & 74.6290* ± 0.9037 & 72.4098* ± 0.6572  \\ 
& JCWM & 66.6154* ± 0.6941 & 66.8379* ± 0.8597 & 62.7267* ± 1.0582 & 68.3410* ± 0.9037 & 64.9100* ± 0.8907  \\ 
& WM-CGM & 59.8562* ± 0.7748 & 59.9640* ± 0.7303 & 55.0599* ± 1.3900 & 59.3251* ± 0.9110 & 57.2308* ± 0.8559  \\ 
& Anomaly & NA & 74.3509* ± 0.9020 & NA & NA & 50.7694* ± 5.7220  \\ 
& NAWM & NA & NA & NA & NA & 72.7749* ± 0.6231  \\ 
FastFOD-Net 
& WM & \textbf{76.6887} ± 0.7102 & \textbf{76.6642} ± 0.9091 & \textbf{70.0695} ± 1.2042 & \textbf{77.3696} ± 1.0330 & \textbf{75.0700} ± 0.7714  \\ 
& JCWM & \textbf{70.3790} ± 0.8130 & \textbf{70.6622} ± 0.9859 & \textbf{64.8251} ± 1.3154 & \textbf{71.8760} ± 1.1395 & \textbf{68.0943} ± 1.0352  \\ 
& WM-CGM & \textbf{61.9567} ± 0.7450 & \textbf{62.1020} ± 0.7746 & \textbf{56.5891} ± 1.3888 & \textbf{61.5614} ± 0.9755 & \textbf{59.8748} ± 0.8822  \\ 
& Anomaly & NA & \textbf{76.6642} ± 0.9091 & NA & NA & \textbf{54.0927} ± 6.1493  \\ 
& NAWM & NA & NA & NA & NA & \textbf{75.4318} ± 0.7438  \\ 

            \toprule
            & & \multicolumn{5}{@{}c@{}}{$\mathbf{r_{Angular}}$\footnotemark[2] ($\uparrow$)} \\\cmidrule{3-7}%
            Methods & Tissue Type & \multicolumn{1}{l}{HCP} & \multicolumn{1}{l}{HCP$_{Anomaly}$} & \multicolumn{1}{l}{Tremor} & \multicolumn{1}{l}{MND} & \multicolumn{1}{l}{MS} \\
            \midrule
SS3T CSD 
& WM & 0.7590* ± 0.0157 & 0.7652* ± 0.0167 & 0.6936* ± 0.0458 & 0.7917* ± 0.0280 & 0.8594* ± 0.0133  \\ 
& JCWM & 0.5265* ± 0.0227 & 0.5367* ± 0.0234 & 0.4960* ± 0.0346 & 0.6091* ± 0.0298 & 0.7535* ± 0.0185  \\ 
& WM-CGM & 0.4534* ± 0.0186 & 0.4554* ± 0.0307 & 0.4744* ± 0.0401 & 0.5755* ± 0.0394 & 0.6930* ± 0.0396  \\ 
& Anomaly & NA & 0.7652* ± 0.0167 & NA & NA & 0.8978* ± 0.0325  \\ 
& NAWM & NA & NA & NA & NA & 0.8442* ± 0.0138  \\ 
FOD-Net 
& WM & 0.8614* ± 0.0124 & 0.8675* ± 0.0117 & 0.7467* ± 0.0519 & 0.8404* ± 0.0247 & 0.8491* ± 0.0141  \\ 
& JCWM & 0.7013* ± 0.0244 & 0.7205* ± 0.0204 & 0.5534* ± 0.0590 & 0.6854* ± 0.0358 & 0.7258* ± 0.0247  \\ 
& WM-CGM & 0.6115* ± 0.0252 & 0.6196* ± 0.0362 & 0.4904* ± 0.0554 & 0.6219* ± 0.0399 & 0.6368* ± 0.0305  \\ 
& Anomaly & NA & 0.8675* ± 0.0117 & NA & NA & 0.8979* ± 0.0296  \\ 
& NAWM & NA & NA & NA & NA & 0.8293* ± 0.0145  \\ 
FastFOD-Net 
& WM & \textbf{0.9155} ± 0.0086 & \textbf{0.9201} ± 0.0072 & \textbf{0.8045} ± 0.0490 & \textbf{0.9042} ± 0.0193 & \textbf{0.9097} ± 0.0108  \\ 
& JCWM & \textbf{0.8286} ± 0.0200 & \textbf{0.8439} ± 0.0130 & \textbf{0.6549} ± 0.0623 & \textbf{0.8147} ± 0.0321 & \textbf{0.8447} ± 0.0191  \\ 
& WM-CGM & \textbf{0.7294} ± 0.0224 & \textbf{0.7372} ± 0.0288 & \textbf{0.5933} ± 0.0577 & \textbf{0.7335} ± 0.0370 & \textbf{0.7738} ± 0.0276  \\ 
& Anomaly & NA & \textbf{0.9201} ± 0.0072 & NA & NA & \textbf{0.9263} ± 0.0303  \\ 
& NAWM & NA & NA & NA & NA & \textbf{0.8979} ± 0.0113  \\
            \bottomrule      

            \end{tabular}
            \begin{tablenotes}
                    \setlength{\itemindent}{0.in}
                    \item $^{3}$MSMT CSD-derived data is considered ground truth (GT). $^{4}$HCP$_{Anomaly}$ denotes HCP healthy subjects with structural abnormalities. $^{5}$WM denotes pure white matter. $^{6}$JCWM and $^{7}$WM-SGM represent the boundaries between WM and juxtacortical or subcortical grey matter.
                    The best result is highlighted in \textbf{bold}, and * denotes results are significantly different from FastFOD-Net ($p < 0.05$). 
            \end{tablenotes}
        \end{threeparttable}
    \end{adjustbox}
\end{table*}

\begin{table*}[t]
    \caption{\textbf{Quantitative analysis of mean angular error$^{1}$ between estimated fixels and MSMT CSD-derived data$^{2}$.}} 
    \label{tab:fod_psnr_ra}
    \centering
    \setlength{\extrarowheight}{1.7pt}
    \begin{adjustbox}{width=\textwidth}
        \begin{threeparttable}
            \begin{tabular}{lllllll}
            \toprule
            & & \multicolumn{5}{@{}c@{}}{$\mathbf{\mu E_{Angular}}$ \footnotemark[1] ($\downarrow$)} \\\cmidrule{3-7}%
            Methods & ROIs\footnotemark[3] & \multicolumn{1}{l}{HCP} & \multicolumn{1}{l}{HCP$_{Anomaly}$} & \multicolumn{1}{l}{Tremor} & \multicolumn{1}{l}{MND} & \multicolumn{1}{l}{MS} \\
            \midrule
SS3T CSD & ROI 1 & 11.3195* ± 0.4641 & 11.4658* ± 0.5575 & 9.9299* ± 0.6195 & 9.3661* ± 0.7447 & 7.1650* ± 0.4200  \\ 
 & ROI 2 & 11.7188* ± 0.7512 & 11.7127* ± 0.8148 & 10.6309* ± 1.0655 & 8.1924* ± 0.9425 & 6.0463* ± 0.8429  \\ 
 & ROI 3 & 14.2819* ± 0.7576 & 14.4520* ± 1.2459 & 14.2531* ± 1.0848 & 9.5959* ± 1.1047 & 8.2231* ± 0.9621  \\ 
 & WM & 14.0733* ± 0.4640 & 13.8267* ± 0.4996 & 15.5042* ± 0.9480 & 12.5016* ± 0.8436 & 9.8694* ± 0.5014  \\ 
 & Anomaly & NA & 13.8267* ± 0.4996 & NA & NA & 6.8722* ± 1.1291  \\ 
 & NAWM & NA & NA & NA & NA & 10.3116* ± 0.4946  \\ 
FOD-Net & ROI 1 & 7.4300* ± 0.4019 & 7.5356* ± 0.4189 & 7.6910* ± 0.5656 & 6.8654* ± 0.5113 & 5.7171* ± 0.2357  \\ 
 & ROI 2 & 7.9312* ± 0.7158 & 7.9235* ± 0.5172 & 8.3075* ± 0.8332 & 6.5644* ± 0.7305 & 6.0729* ± 0.5926  \\ 
 & ROI 3 & 8.7554* ± 0.7193 & 9.0422* ± 1.1177 & 11.3200* ± 1.1015 & 8.2244* ± 0.9652 & 8.0804* ± 0.9460  \\ 
 & WM & 10.5850* ± 0.4406 & 10.3668* ± 0.4374 & 13.4913* ± 1.2522 & 10.3761* ± 0.6715 & 9.2633* ± 0.4295  \\ 
 & Anomaly & NA & 10.3668* ± 0.4374 & NA & NA & 6.9770* ± 0.8948  \\ 
 & NAWM & NA & NA & NA & NA & 9.7037* ± 0.4174  \\ 
FastFOD-Net & ROI 1 & \textbf{4.1873} ± 0.2889 & \textbf{4.3173} ± 0.2846 & \textbf{5.2427} ± 0.4955 & \textbf{4.0590} ± 0.3721 & \textbf{3.7379} ± 0.2065  \\ 
 & ROI 2 & \textbf{6.0944} ± 0.5396 & \textbf{6.0960} ± 0.4258 & \textbf{6.9762} ± 0.8086 & \textbf{4.8787} ± 0.4501 & \textbf{4.5656} ± 0.4558  \\ 
 & ROI 3 & \textbf{6.4689} ± 0.6401 & \textbf{6.8098} ± 0.5127 & \textbf{9.2458} ± 1.0695 & \textbf{5.8457} ± 0.5932 & \textbf{6.1278} ± 0.7135  \\ 
 & WM & \textbf{7.6581} ± 0.3529 & \textbf{7.5290} ± 0.3278 & \textbf{11.3303} ± 1.3980 & \textbf{7.7757} ± 0.6271 & \textbf{7.1719} ± 0.4620  \\ 
 & Anomaly & NA & \textbf{7.5290} ± 0.3278 & NA & NA & \textbf{4.9169} ± 0.8623  \\ 
 & NAWM & NA & NA & NA & NA & \textbf{7.5016} ± 0.4554  \\ 
            \bottomrule      

            \end{tabular}
            \begin{tablenotes}
                    \setlength{\itemindent}{0.in}
                \item $^{2}$ MSMT CSD-derived data is considered ground truth (GT).
                $^{3}$ The regions labeled as ROI 1, 2, and 3 represent areas with distinct fiber configurations, each occurring with a varying number of fixels (1, 2, and 3, respectively); WM denotes white matter. Anomaly refers to ROIs surrounding the anomalous regions on the HCP$_{Anomaly}$ cases, and lesions on MS cases, respectively. NAWM represents normal-appearing WM outside lesions. 
                The best result is highlighted in \textbf{bold}, and * indicates results are significantly different from FastFOD-Net ($p < 0.05$).
            \end{tablenotes}
        \end{threeparttable}
    \end{adjustbox}
\end{table*}


\begin{table*}[t]
    \caption{\textbf{Quantitative analysis of FD error$^{1}$ of fixels between estimates fixels and MSMT CSD-derived data$^{2}$.}} 
    \label{tab:fod_psnr_ra}
    \centering
    \setlength{\extrarowheight}{1.7pt}
    \begin{adjustbox}{width=\textwidth}
        \begin{threeparttable}
            \begin{tabular}{lllllll}
            \toprule
            & & \multicolumn{5}{@{}c@{}}{$\mathbf{E_{FD}}$\footnotemark[1] ($\downarrow$)} \\\cmidrule{3-7}%
            Methods & ROIs\footnotemark[3] & \multicolumn{1}{l}{HCP} & \multicolumn{1}{l}{HCP$_{Anomaly}$$^{4}$} & \multicolumn{1}{l}{Tremor} & \multicolumn{1}{l}{MND} & \multicolumn{1}{l}{MS} \\
            \midrule
SS3T CSD & ROI 1 & 0.3964* ± 0.0183 & 0.3964* ± 0.0212 & 0.3023* ± 0.0321 & 0.2849* ± 0.0369 & 0.2099* ± 0.0154  \\ 
 & ROI 2 & 0.2644* ± 0.0211 & 0.2642* ± 0.0282 & 0.2476* ± 0.0356 & 0.1978* ± 0.0348 & 0.1280* ± 0.0216  \\ 
 & ROI 3 & 0.4206* ± 0.0225 & 0.4091* ± 0.0236 & 0.4641* ± 0.0392 & 0.3205* ± 0.0413 & 0.2605* ± 0.0305  \\ 
 & WM & 0.3955* ± 0.0161 & 0.3881* ± 0.0169 & 0.4203* ± 0.0371 & 0.3221* ± 0.0291 & 0.2437* ± 0.0138  \\ 
 & Anomaly & NA & 0.3881* ± 0.0169 & NA & NA & 0.2759* ± 0.0469  \\ 
 & NAWM & NA & NA & NA & NA & 0.2500* ± 0.0135  \\ 
FOD-Net & ROI 1 & 0.1668* ± 0.0117 & 0.1668* ± 0.0086 & 0.1616* ± 0.0215 & 0.1408* ± 0.0140 & 0.1240* ± 0.0104  \\ 
 & ROI 2 & 0.1939* ± 0.0183 & 0.1884* ± 0.0131 & 0.2444* ± 0.0292 & 0.1724* ± 0.0298 & 0.1585* ± 0.0210  \\ 
 & ROI 3 & 0.3151* ± 0.0194 & 0.3123* ± 0.0233 & 0.4235* ± 0.0375 & 0.3312* ± 0.0351 & 0.3303* ± 0.0326  \\ 
 & WM & 0.2495* ± 0.0126 & 0.2401* ± 0.0080 & \textbf{0.3428*} ± 0.0448 & 0.2363* ± 0.0180 & 0.2171* ± 0.0178  \\ 
 & Anomaly & NA & 0.2401* ± 0.0080 & NA & NA & 0.2414* ± 0.0373  \\ 
 & NAWM & NA & NA & NA & NA & 0.2207* ± 0.0177  \\ 
FastFOD-Net & ROI 1 & \textbf{0.0845} ± 0.0049 & \textbf{0.0898} ± 0.0087 & \textbf{0.1029} ± 0.0163 & \textbf{0.0786} ± 0.0107 & \textbf{0.0728} ± 0.0087  \\ 
 & ROI 2 & \textbf{0.1471} ± 0.0144 & \textbf{0.1464} ± 0.0107 & \textbf{0.1707} ± 0.0263 & \textbf{0.1208} ± 0.0237 & \textbf{0.1076} ± 0.0187  \\ 
 & ROI 3 & \textbf{0.2761} ± 0.0228 & \textbf{0.2828} ± 0.0213 & \textbf{0.3660} ± 0.0365 & \textbf{0.2409} ± 0.0303 & \textbf{0.2421} ± 0.0250  \\ 
 & WM & \textbf{0.2177} ± 0.0119 & \textbf{0.2088} ± 0.0092 & 0.3491 ± 0.0603 & \textbf{0.2027} ± 0.0217 & \textbf{0.1733} ± 0.0181  \\ 
 & Anomaly & NA & \textbf{0.2088} ± 0.0092 & NA & NA & \textbf{0.1654} ± 0.0336  \\ 
 & NAWM & NA & NA & NA & NA & \textbf{0.1787} ± 0.0178  \\ 
            \bottomrule      

            \end{tabular}
            \begin{tablenotes}
                    \setlength{\itemindent}{0.in}
                \item $^{2}$ MSMT CSD-derived data is considered ground truth (GT).
                $^{3}$ The regions labeled as ROI 1, 2, and 3 represent areas with distinct fiber configurations, each occurring with a varying number of fixels (1, 2, and 3, respectively); WM denotes white matter. Anomaly refers to ROIs surrounding the anomalous regions on the HCP$_{Anomaly}$ cases, and lesions on MS cases, respectively. NAWM represents normal-appearing WM outside lesions. 
                The best result is highlighted in \textbf{bold}, and * indicates results are significantly different from FastFOD-Net ($p < 0.05$).
            \end{tablenotes}
        \end{threeparttable}
    \end{adjustbox}
\end{table*}

\begin{table*}[t]
    \caption{\textbf{Quantitative analysis of peak error$^{1}$ of fixels between estimated fixels and multi-shell HARDI data$^{2}$.}} 
    \label{tab:fod_psnr_ra}
    \centering
    \setlength{\extrarowheight}{1.7pt}
    \begin{adjustbox}{width=\textwidth}
        \begin{threeparttable}
            \begin{tabular}{lllllll}
            \toprule
            & & \multicolumn{5}{@{}c@{}}{$\mathbf{E_{Peak}}$\footnotemark[1] ($\downarrow$)} \\\cmidrule{3-7}%
            Methods & ROIs\footnotemark[3] & \multicolumn{1}{l}{HCP} & \multicolumn{1}{l}{HCP$_{Anomaly}$$^{4}$} & \multicolumn{1}{l}{Tremor} & \multicolumn{1}{l}{MND} & \multicolumn{1}{l}{MS} \\
            \midrule
SS3T CSD & ROI 1 & 0.3996* ± 0.0184 & 0.4036* ± 0.0218 & 0.3159* ± 0.0348 & 0.2827* ± 0.0416 & 0.2027* ± 0.0149  \\ 
 & ROI 2 & 0.3211* ± 0.0189 & 0.3219* ± 0.0254 & 0.2787 ± 0.0362 & 0.2489* ± 0.0369 & \textbf{0.1644*} ± 0.0231  \\ 
 & ROI 3 & 0.4832* ± 0.0226 & 0.4701* ± 0.0254 & 0.5225* ± 0.0416 & 0.3710* ± 0.0449 & \textbf{0.2964*} ± 0.0311  \\ 
 & WM & 0.4165* ± 0.0179 & 0.4095* ± 0.0193 & 0.4742* ± 0.0501 & 0.3411* ± 0.0346 & 0.2566* ± 0.0158  \\ 
 & Anomaly & NA & 0.4095* ± 0.0193 & NA & NA & 0.2484* ± 0.0399  \\ 
 & NAWM & NA & NA & NA & NA & 0.2655* ± 0.0156  \\ 
FOD-Net & ROI 1 & 0.1455* ± 0.0111 & 0.1453* ± 0.0073 & 0.1560* ± 0.0208 & 0.1322* ± 0.0151 & 0.1152* ± 0.0086  \\ 
 & ROI 2 & \textbf{0.2581*} ± 0.0246 & \textbf{0.2475} ± 0.0186 & \textbf{0.2734*} ± 0.0262 & 0.2241* ± 0.0384 & 0.2042* ± 0.0268  \\ 
 & ROI 3 & 0.3940 ± 0.0254 & 0.3839 ± 0.0276 & 0.5074* ± 0.0498 & 0.4112* ± 0.0484 & 0.4166* ± 0.0461  \\ 
 & WM & 0.2408* ± 0.0155 & 0.2279* ± 0.0091 & \textbf{0.3888} ± 0.0657 & 0.2480* ± 0.0231 & 0.2380* ± 0.0253  \\ 
 & Anomaly & NA & 0.2279* ± 0.0091 & NA & NA & 0.2196* ± 0.0321  \\ 
 & NAWM & NA & NA & NA & NA & 0.2450* ± 0.0250  \\ 
FastFOD-Net & ROI 1 & \textbf{0.0906} ± 0.0075 & \textbf{0.0863} ± 0.0065 & \textbf{0.1126} ± 0.0121 & \textbf{0.0805} ± 0.0115 & \textbf{0.0768} ± 0.0074  \\ 
 & ROI 2 & 0.2688 ± 0.0287 & 0.2497 ± 0.0275 & 0.2887 ± 0.0373 & \textbf{0.2089} ± 0.0413 & 0.1841 ± 0.0282  \\ 
 & ROI 3 & \textbf{0.3847} ± 0.0342 & \textbf{0.3814} ± 0.0363 & \textbf{0.4603} ± 0.0494 & \textbf{0.3263} ± 0.0511 & 0.3394 ± 0.0314  \\ 
 & WM & \textbf{0.2161} ± 0.0181 & \textbf{0.1974} ± 0.0163 & 0.3929 ± 0.0784 & \textbf{0.2140} ± 0.0319 & \textbf{0.1970} ± 0.0283  \\ 
 & Anomaly & NA & \textbf{0.1974} ± 0.0163 & NA & NA & \textbf{0.1315} ± 0.0318  \\ 
 & NAWM & NA & NA & NA & NA & \textbf{0.2046} ± 0.0277  \\ 
            \bottomrule      

            \end{tabular}
            \begin{tablenotes}
                    \setlength{\itemindent}{0.in}
                    \item $^{3}$Multi-shell HARDI data is considered ground truth (GT). $^{4}$HCP$_{Anomaly}$ denotes HCP healthy subjects with structural abnormalities. $^{5}$WM denotes pure white matter. $^{6}$JCWM and $^{7}$WM-SGM represent the boundaries between WM and juxtacortical or subcortical grey matter. The best result is highlighted in \textbf{bold}, and * denotes results are significantly different from FastFOD-Net ($p < 0.05$). 
            \end{tablenotes}
        \end{threeparttable}
    \end{adjustbox}
\end{table*}

\begin{table*}[t]
    \caption{\textbf{Quantitative analysis of connectome assessment between MSMT CSD-derived connectomes$^{1}$ and other LARDI-based estimates.}}\label{tab:connectome_metrics}
        \centering
        \setlength\tabcolsep{0.6em}
        \begin{adjustbox}{width=\textwidth}
        \begin{threeparttable}
        \begin{tabular}{llllll}
            \toprule
            & \multicolumn{5}{@{}c@{}}{$\tau$\footnotemark[2] ($\uparrow$)} \\\cmidrule{2-6}%
            Methods & \multicolumn{1}{l}{HCP} & \multicolumn{1}{l}{HCP$_{Anomaly}$\footnotemark[3]} & \multicolumn{1}{l}{Tremor} & \multicolumn{1}{l}{MND} & \multicolumn{1}{l}{MS} \\
            \midrule
            SS3T CSD & 0.8484 ± 0.0075 & 0.8590 ± 0.0092 & 0.8739 ± 0.0179 & 0.8551 ± 0.0149 & 0.8672 ± 0.0237 \\
            FOD-Net & 0.8991 ± 0.0063 & 0.9029 ± 0.0079 & 0.7415 ± 0.0198 & 0.7481 ± 0.0188 & 0.8673 ± 0.0227 \\
            FastFOD-Net & \textbf{0.9169} ± 0.0058 & \textbf{0.9186} ± 0.0059 & \textbf{0.9050} ± 0.0132 & \textbf{0.9100} ± 0.0125 & \textbf{0.9018} ± 0.0139 \\

            \toprule
            & \multicolumn{5}{@{}c@{}}{$\mathbf{\mu{Disparity}}$\footnotemark[4] ($\downarrow$)} \\\cmidrule{2-6}%
            Methods & \multicolumn{1}{l}{HCP} & \multicolumn{1}{l}{HCP$_{Anomaly}$} & \multicolumn{1}{l}{Tremor} & \multicolumn{1}{l}{MND} & \multicolumn{1}{l}{MS} \\
            \midrule
            SS3T CSD & 553.1383 & 488.6859 & 444.9234 & 380.1648 & 357.9414 \\
            FOD-Net & 261.0568 & 230.0609 & 506.5477 & 573.3384 & 414.1002 \\
            FastFOD-Net & \textbf{218.0480} & \textbf{188.9277} & \textbf{219.3831} & \textbf{336.5554} & \textbf{186.4356} \\

            \toprule
            & \multicolumn{5}{@{}c@{}}{$\mathbf{Significant}\ \mathbf{Edge}\ \mathbf{Difference}$ \footnotemark[5] ($\downarrow$)} 
            \\\cmidrule{2-6}%
            Methods & \multicolumn{1}{l}{HCP} & \multicolumn{1}{l}{HCP$_{Anomaly}$$^{4}$} & \multicolumn{1}{l}{Tremor} & \multicolumn{1}{l}{MND} & \multicolumn{1}{l}{MS} \\
            \midrule
            SS3T CSD & 50.57\% & 36.49\% & 48.71\% & 68.42\% & 54.84\% \\
            FOD-Net & 24.04\% & 6.83\% & 73.55\% & 75.96\% & 50.08\%\\
            FastFOD-Net & \textbf{22.69\%} & \textbf{1.81\%} & \textbf{30.01\%} & \textbf{32.53\%} & \textbf{38.49\%} \\
            \bottomrule
        \end{tabular}
        \begin{tablenotes}
                    \setlength{\itemindent}{0.in}
                    \item $^{1}$Connectome via MSMT CSD is considered the ground truth (GT). $^{3}$HCP$_{Anomaly}$ denotes HCP healthy subjects with structural abnormalities. $^{2}$$\tau$ denotes mean Kendall ranking correlation coefficient, and $^{4}$$\mathbf{\mu{Disparity}}$ denotes the mean disparity across all cases. $^{5}$The number of significantly different edges (determined by edge-wise paired t-tests with $p < 0.05$) is presented in percentage (\%). The best result is highlighted in \textbf{bold}.
        \end{tablenotes}
        \end{threeparttable}
        \end{adjustbox}
\end{table*}
\clearpage

\begin{table*}[t]
    \caption{\textbf{Quantitative analysis of graph metrics$^{1}$ of connectomes between estimates and multi-shell HARDI data$^{2}$.}} 
    \label{tab:fod_psnr_ra}
    \centering
    \setlength{\extrarowheight}{1.7pt}
    \begin{adjustbox}{width=\textwidth}
        \begin{threeparttable}
            \begin{tabular}{llrrrrr}
            \toprule
            Methods & Graph Metrics\footnotemark[3] & \multicolumn{1}{l}{HCP} & \multicolumn{1}{l}{HCP$_{Anomaly}$$^{4}$} & \multicolumn{1}{l}{Tremor} & \multicolumn{1}{l}{MND} & \multicolumn{1}{l}{MS} \\
            \midrule
SS3T CSD~\cite{dhollander2016unsupervised} 
&   Assortativity & -0.0220 & -0.0224 & -0.0273 & -0.0204 & -0.0307 \\
& DR$^{5}$ & 13.04\% & 16.46\% & 8.38\% & 8.18\% & 4.36\% \\
FOD-Net~\cite{zeng2022fod} 
&   Assortativity & -0.0220 & -0.0227 & -0.0252 & -0.0258 & -0.0303 \\
& DR & 12.49\% & 16.85\% & 2.80\% & 36.93\% & 2.02\% \\
FastFOD-Net 
&   Assortativity & -0.0196 & -0.0199 & -0.0259 & -0.0194 & -0.0297 \\
& DR & -0.51\% & 2.29\% & 2.47\% & 2.01\% & -0.17\% \\
MSMT CSD~\cite{jeurissen2014multi}
&   Assortativity & -0.0197 & -0.0194 & -0.0256 & -0.0190 & -0.0301 \\
& DR & - & - & - & - & - \\
\hline
SS3T CSD 
& Density & 0.8469 & 0.8474 & 0.8387 & 0.8487 & 0.8194 \\
& DR & -0.29\% & 0.08\% & -0.86\% & -0.19\% & 1.49\% \\
FOD-Net 
& Density & 0.8436 & 0.8397 & 0.6705 & 0.7312 & 0.8032 \\
& DR & -0.67\% & -0.84\% & -20.76\% & -14.01\% & -0.53\% \\
FastFOD-Net 
& Density & 0.8463 & 0.8449 & 0.8372 & 0.8470 & 0.8055 \\
& DR & -0.36\% & -0.22\% & -1.05\% & -0.39\% & -0.26\% \\
MSMT CSD 
& Density & 0.8494 & 0.8468 & 0.8461 & 0.8503 & 0.8082 \\
& DR & - & - & - & - & - \\
\hline
SS3T CSD 
& Global Efficiency & 4447.0388 & 4464.7697 & 4319.4735 & 4182.7957 & 4400.0063 \\
& DR & -20.08\% & -16.12\% & -11.39\% & -14.12\% & -9.29\% \\
FOD-Net 
& Global Efficiency & 5338.0550 & 5411.4844 & 4484.0406 & 4627.5225 & 5608.3059 \\
& DR & -4.10\% & 1.68\% & -8.12\% & -5.01\% & 15.71\% \\
FastFOD-Net 
& Global Efficiency & 5219.9786 & 5265.3119 & 5180.2524 & 5132.2079 & 5053.7391 \\
& DR & -6.20\% & -1.06\% & 6.10\% & 5.43\% & 4.22\% \\
MSMT CSD 
& Global Efficiency & 5566.7768 & 5323.2690 & 4882.6422 & 4873.7886 & 4851.8529 \\
& DR & - & - & - & - & - \\
\hline
SS3T CSD & Mean Strength & 101702.3121 & 101833.3209 & 102284.3668 & 95626.6352 & 98913.0292 \\
 & DR & -20.61\% & -15.95\% & -11.59\% & -14.40\% & -8.60\% \\ 
FOD-Net & Mean Strength & 121284.3588 & 120383.1025 & 101168.4562 & 104222.5172 & 124110.1985 \\
 & DR & -5.33\% & -0.62\% & -12.58\% & -6.69\% & 14.77\% \\ 
FastFOD-Net & Mean Strength & 118756.9353 & 117726.1317 & 117652.9900 & 116287.9850 & 111721.8927 \\
 & DR & -7.30\% & -2.81\% & 1.63\% & 4.14\% & 3.24\% \\ 
MSMT CSD & Mean Strength & 128106.0299 & 121139.1854 & 115762.0262 & 111809.4013 & 108339.0105 \\
 & DR & - & - & - & - & - \\ 
\hline
SS3T CSD & Transitivity & 400.9858 & 390.9809 & 381.3502 & 368.1949 & 366.7435 \\
 & DR & -26.56\% & -23.35\% & -17.80\% & -22.28\% & -14.07\% \\ 
FOD-Net & Transitivity & 510.6750 & 492.3965 & 434.2220 & 440.5912 & 514.7142 \\
 & DR & -6.47\% & -3.48\% & -6.10\% & -6.75\% & 20.82\% \\ 
FastFOD-Net & Transitivity & 499.1105 & 480.6893 & 468.3477 & 486.3475 & 440.6813 \\
 & DR & -8.58\% & -5.74\% & 0.81\% & 2.86\% & 3.14\% \\ 
MSMT CSD & Transitivity & 545.9997 & 510.1657 & 465.3202 & 474.3285 & 428.6459 \\
 & DR & - & - & - & - & - \\ 
\bottomrule
            \end{tabular}
            \begin{tablenotes}
                    \setlength{\itemindent}{0.in}
                    \item $^{3}$Multi-shell HARDI data is considered the ground truth (GT). $^{4}$HCP$_{Anomaly}$ denotes HCP healthy subjects with structural abnormalities. For each method, $^{5}$DR is the difference ratio of graph metrics when compared to GT. 
            \end{tablenotes}
        \end{threeparttable}
    \end{adjustbox}
\end{table*}
\clearpage

\begin{table*}[t]
    \caption{\textbf{Quantitative analysis of graph metrics$^{1}$ of connectomes between estimates and multi-shell HARDI data$^{2}$}.} 
    \label{tab:fod_psnr_ra}
    \centering
    \setlength{\extrarowheight}{1.7pt}
    \begin{adjustbox}{width=\textwidth}
        \begin{threeparttable}
            \begin{tabular}{llrrrrr}
            \toprule
            Methods & Graph Metrics\footnotemark[3] & \multicolumn{1}{l}{HCP} & \multicolumn{1}{l}{HCP$_{Anomaly}$$^{4}$} & \multicolumn{1}{l}{Tremor} & \multicolumn{1}{l}{MND} & \multicolumn{1}{l}{MS} \\
            \midrule
SS3T CSD~\cite{dhollander2016unsupervised} & Mean Betweeness Centrality & 153.2595 & 156.2397 & 154.3462 & 156.5223 & 156.2325 \\
 & DR$^{5}$ & 1.58\% & 5.21\% & 0.78\% & 0.69\% & 2.53\% \\
FOD-Net~\cite{zeng2022fod} & Mean Betweeness Centrality & 152.4310 & 154.1476 & 141.3419 & 139.2754 & 154.8807 \\
 & DR & 1.00\% & 3.79\% & -7.70\% & -10.43\% & 1.67\% \\
FastFOD-Net & Mean Betweeness Centrality & 147.0679 & 151.9794 & 152.6960 & 153.1659 & 150.5734 \\
 & DR & -2.53\% & 2.26\% & -0.35\% & -1.51\% & -1.22\% \\
MSMT CSD~\cite{jeurissen2014multi} & Mean Betweeness Centrality & 151.7024 & 149.2206 & 153.5751 & 155.8800 & 152.6637 \\
 & DR & - & - & - & - & - \\
\hline
SS3T CSD & Mean Clustering Coefficient & 366.9251 & 358.0020 & 346.6936 & 337.7492 & 330.1410 \\
 & DR & -26.63\% & -23.28\% & -18.07\% & -22.26\% & -13.34\% \\
FOD-Net & Mean Clustering Coefficient & 465.1069 & 447.4440 & 368.3558 & 377.9732 & 457.1846 \\
 & DR & -7.00\% & -4.15\% & -12.58\% & -12.77\% & 20.21\% \\
FastFOD-Net & Mean Clustering Coefficient & 456.2180 & 439.1776 & 424.1803 & 444.6616 & 392.0182 \\
 & DR & -8.78\% & -5.87\% & 0.11\% & 2.54\% & 2.76\% \\
MSMT CSD & Mean Clustering Coefficient & 500.1260 & 466.7638 & 424.4369 & 435.0083 & 382.8577 \\
 & DR & - & - & - & - & - \\
\hline
SS3T CSD & Mean Local Efficiency & 459.7633 & 450.6862 & 431.9841 & 424.8624 & 423.3546 \\
 & DR & -27.28\% & -23.77\% & -18.34\% & -22.37\% & -14.11\% \\
FOD-Net & Mean Local Efficiency & 589.9269 & 572.6218 & 474.5110 & 488.2660 & 592.3535 \\
 & DR & -6.69\% & -3.15\% & -9.99\% & -10.60\% & 20.35\% \\
FastFOD-Net & Mean Local Efficiency & 575.9728 & 558.6157 & 539.5586 & 563.3202 & 508.4929 \\
 & DR & -8.90\% & -5.50\% & 1.87\% & 3.10\% & 3.09\% \\
MSMT CSD & Mean Local Efficiency & 632.2549 & 591.2551 & 530.4742 & 547.8706 & 494.5935 \\
 & DR & - & - & - & - & - \\
\hline   
SS3T CSD & Mean Modularity & 3.1036 & 3.0698 & 2.7955 & 3.1372 & 3.2706 \\
 & DR & -3.67\% & -4.81\% & -0.78\% & -2.03\% & -0.96\% \\
FOD-Net & Mean Modularity & 3.1256 & 3.1111 & 2.8303 & 2.8271 & 2.9195 \\
 & DR & -3.11\% & -4.56\% & 0.30\% & -11.53\% & -12.61\% \\
FastFOD-Net & Mean Modularity & 3.2827 & 3.2611 & 2.9374 & 3.3290 & 3.5195 \\
 & DR & 2.53\% & 0.77\% & 4.11\% & 3.51\% & 4.75\% \\
MSMT CSD & Mean Modularity & 3.2613 & 3.3198 & 2.9292 & 3.2718 & 3.4462 \\
 & DR & - & - & - & - & - \\
\hline
SS3T CSD & Mean Network Degree & 70.2893 & 70.3317 & 69.6160 & 70.4396 & 68.0142 \\
 & DR & -0.29\% & 0.08\% & -0.86\% & -0.19\% & 1.49\% \\ 
FOD-Net & Mean Network Degree & 70.0214 & 69.6952 & 55.6538 & 60.6919 & 66.6677 \\
 & DR & -0.67\% & -0.84\% & -20.76\% & -14.01\% & -0.53\% \\ 
FastFOD-Net & Mean Network Degree & 70.2429 & 70.1302 & 69.4841 & 70.2997 & 66.8604 \\
 & DR & -0.36\% & -0.22\% & -1.05\% & -0.39\% & -0.26\% \\ 
MSMT CSD & Mean Network Degree & 70.4964 & 70.2825 & 70.2253 & 70.5729 & 67.0800 \\
 & DR & - & - & - & - & - \\ 
\bottomrule
            \end{tabular}
            \begin{tablenotes}
                    \setlength{\itemindent}{0.in}
                    \item $^{3}$Multi-shell HARDI data is considered the ground truth (GT). $^{4}$HCP$_{Anomaly}$ denotes HCP healthy subjects with structural abnormalities. For each method, $^{5}$DR is the difference ratio of graph metrics when compared to GT. 
            \end{tablenotes}
        \end{threeparttable}
    \end{adjustbox}
\end{table*}
\clearpage

\clearpage

\begin{table*}[t]
    \caption{\textbf{Quantitative comparisons of significantly different fixels between the fiber density (FD) of amyotrophic lateral sclerosis (ALS) patients and healthy subjects.} In this comparison, significant fixels derived from the reference method (MSMT CSD) were juxtaposed with those obtained from alternative methodologies (specifically, SS3T CSD, FOD-Net, and FastFOD-Net). }
    \label{tab:quantitative_fba}
    \centering
    \begin{tabular}{{l c c c c}}
        \toprule
        & \multicolumn{4}{@{}c@{}}{CC\footnotemark[2]} \\\cmidrule{2-5}%
        \multicolumn{1}{c}{} 
        & $^{1}$\textbf{Sensitivity} ($\uparrow$)
        & $^{2}$\textbf{Specificity} ($\uparrow$)
        & $^{3}$\textbf{Precision} ($\uparrow$)
        & $^{4}$\textbf{F1} ($\uparrow$)\\
        \midrule
        SS3T CSD 
        & 0.4662 & \textbf{0.9708} & 0.9385 & 0.6229 \\
        FOD-Net 
        & 0.4659 & 0.9698 & 0.9390 & 0.6228 \\
        FastFOD-Net
        & \textbf{0.7014} & 0.9586 & \textbf{0.9445} & \textbf{0.8050} \\
        
        \toprule
        & \multicolumn{4}{@{}c@{}}{CST Right\footnotemark[2]} \\\cmidrule{2-5}%
        \multicolumn{1}{c}{} 
        & $^{1}$\textbf{Sensitivity} ($\uparrow$)
        & $^{2}$\textbf{Specificity} ($\uparrow$)
        & $^{3}$\textbf{Precision} ($\uparrow$)
        & $^{4}$\textbf{F1} ($\uparrow$)\\
        \midrule
        SS3T CSD 
        & 0.3676 & 0.9653 & 0.9195 & 0.5253 \\
        FOD-Net 
        & 0.4320 & 0.9419 & 0.8939 & 0.5825 \\
        FastFOD-Net
        & \textbf{0.4575} & \textbf{0.9742} & \textbf{0.9560} & \textbf{0.6188} \\
        
        \toprule
        & \multicolumn{4}{@{}c@{}}{CST Left\footnotemark[2]} \\\cmidrule{2-5}%
        \multicolumn{1}{c}{} 
        & $^{1}$\textbf{Sensitivity} ($\uparrow$)
        & $^{2}$\textbf{Specificity} ($\uparrow$)
        & $^{3}$\textbf{Precision} ($\uparrow$)
        & $^{4}$\textbf{F1} ($\uparrow$)\\
        \midrule
        SS3T CSD 
        & 0.3604 & \textbf{0.9823} & \textbf{0.9599} & 0.5240 \\
        FOD-Net 
        & \textbf{0.4565} & 0.9276 & 0.8891 & \textbf{0.6032} \\
        FastFOD-Net
        & 0.4256 & 0.9569 & 0.9292 & 0.5838 \\
        \bottomrule
    \end{tabular}
\end{table*}

\begin{table*}[t]
\caption{\textbf{ANOVA analysis of pathology-relevant thalamic connections across tremor disorders.} Statistical comparison of thalamic interconnections between tremor disorders, i.e., Parkinson’s disease (PD), essential tremor (ET), and dystonic tremor (DT) across different methods. }
\label{tab:anova_tremor}
\centering
\begin{threeparttable}
\begin{tabular}{ll r r c c r}
\toprule
Method & Factor & DF$^{1}$ & SS$^{2}$ & Mean Square$^{3}$ & $p$ \\ \midrule 
SS3T CSD & Between (Group) & 2 & 23.8401 & 11.9201 & 0.5904  \\ 
& Within (Residual) & 36 & 802.4890  & 22.2913 &  &  \\ 
FOD-Net & Between & 2 & 0.2355  & 0.1178 & 0.8488  \\
& Within & 36 & 25.7489  & 0.7152 &  &   \\ 
FastFOD-Net & Between & 2 & 7.7358  & 3.8679 & 0.6598 \\ 
& Within & 36 & 330.9880  & 9.1941 &  &  \\
MSMT CSD & Between & 2 & 9.0563 & 4.5282 & 0.5195 \\ 
& Within & 36 & 244.4170 & 6.7894 &  &  \\
\bottomrule
\end{tabular}
\begin{tablenotes}
    \setlength{\itemindent}{0.in}
    \item $^{1}$DF is the degrees of freedom associated with the factor. $^{2}$Sum of squares (SS) and $^{3}$mean square represents the total variability and the average variation due to the factor respectively.   
\end{tablenotes}
\end{threeparttable}
\end{table*}
\clearpage

\end{document}